\documentclass[
reprint,
raggedbottom
]{revtex4-2}
\usepackage[T1]{fontenc}
\usepackage[utf8]{inputenc}

\usepackage{amsmath}
\usepackage{amssymb}
\usepackage{calc}
\usepackage{graphicx}
\usepackage{xcolor}
\usepackage{siunitx}\DeclareSIUnit\angstrom{\text{Å}}
\usepackage[hidelinks]{hyperref}
\usepackage{algorithm}
\usepackage{booktabs}
\usepackage{algpseudocode}
\usepackage{setspace}
\usepackage{gensymb}
\usepackage{soul}
\usepackage{mhchem}
\usepackage{bbm}
\DeclareMathOperator{\E}{\mathbbm{E}}

\newcommand*{\tran}{\mathsf{T}}

\newcommand{\tmm}{token-merging module}
\newcommand{\TMM}{Token-Merging Module (TMM)}
\newcommand{\fgc}{fragment-graph convolution}
\newcommand{\wmlp}{windowed MLP}

\newcommand{\fragtoken}{fragment-level token}
\newcommand{\tokenmixer}{transformer token mixer}
\newcommand{\Transformer}{transformer}
\begin{document}

\title{Hierarchical geometric deep learning enables scalable analysis of molecular dynamics}

\author{Zihan Pengmei}
\affiliation{Department of Chemistry and James Franck Institute, University of Chicago, Chicago, Illinois 60637, United States}
\author{Spencer C. Guo}
\affiliation{Department of Chemistry and James Franck Institute, University of Chicago, Chicago, Illinois 60637, United States}
\author{Chatipat Lorpaiboon}
\affiliation{Department of Chemistry and James Franck Institute, University of Chicago, Chicago, Illinois 60637, United States}
%\author{Jonathan Weare}
%\affiliation{Courant Institute of Mathematical Sciences, New York University, New York, New York 10012, United States}
\author{Aaron R. Dinner}
\email{dinner@uchicago.edu}
\affiliation{Department of Chemistry and James Franck Institute, University of Chicago, Chicago, Illinois 60637, United States}

\begin{abstract}
Molecular dynamics simulations can generate atomically detailed trajectories of complex systems, but analyzing these dynamics can be challenging when systems lack well-established quantitative descriptors (features).  Graph neural networks (GNNs) in which messages are passed between nodes that represent atoms that are spatial neighbors promise to obviate manual feature engineering, but the use of GNNs with biomolecular systems of more than a few hundred residues has been limited in the context of analyzing dynamics by both difficulties in capturing the details of long-range interactions with message passing and the memory and runtime requirements associated with large graphs. Here, we show how local information can be aggregated to reduce memory and runtime requirements without sacrificing atomic detail.  We demonstrate that this approach opens the door to analyzing simulations of protein-nucleic acid complexes with thousands of residues on single GPUs within minutes. For systems with hundreds of residues, for which there are sufficient data to make quantitative comparisons, we show that the approach improves performance and interpretability.

%Understanding how biomolecules interact and adopt diverse conformations is essential for elucidating their functions and informing drug discovery. Molecular dynamics simulations can generate atomically detailed trajectories of large, heterogeneous complexes. However, analyzing systems with $10^3$ to $10^4$ components (such as amino acids, ligands, and nucleic acids) at all-atom resolution remains prohibitively expensive. Conventional all-atom graph neural networks (GNNs) lack long-range expressivity and struggle to scale beyond tens of biological fragments, limiting their utility for large-scale mechanistic insights. Here, we introduce an auxiliary token merging module for the previously introduced geom2vec framework that merges local structural ``tokens'' into higher-order tokens (``fragments'') without sacrificing atomic detail. This token-merging approach dramatically reduces memory and runtime requirements, enabling us to analyze molecular dynamics simulations of massive biomolecular assemblies on a single GPU within minutes. As a demonstration, we showcase the ability of the network to identify metastable states of a 592-residue SARS-CoV-2 helicase complex. By scaling expensive geometric deep learning methods to previously inaccessible biomolecular systems, our framework provides a powerful tool for studying complex conformational changes that underlie biological function.
\end{abstract}

\maketitle

\section{Introduction}

Elucidating how biomolecules interconvert among their conformations is essential for understanding how they function and in turn designing therapies and harnessing them for engineering applications. Advances in experimental techniques such as cryogenic electron microscopy (cryo-EM) now enable near-atomic resolution imaging of diverse conformations of large biomolecular complexes\cite{singer_computational_2020, tang_conformational_2023}, and advances in computing enable simulating the dynamics of such systems \cite{shaw_anton_2014, shaw_anton_2021,shirts_computing_2000, zimmerman_sarscov2_2021}. However, interpreting these data remains challenging owing to their high dimensionality.  
While many machine-learning approaches for identifying low-dimensional representations that capture key dynamical processes have been introduced \cite{ma_automatic_2005, rohrdanz_determination_2011, mardt_vampnets_2018, wang2021state, chen_discovering_2023, bonati2023unified,zhang2024descriptor, zou2025graph}, they often still rely on manual feature engineering.

Because biomolecular structures can be represented as three-dimensional point clouds, geometric graph neural networks (GNNs) are naturally suited to learning molecular features~\cite{gilmer2017neural,jamasb_evaluating_2023,xie2019graph, patel2024graph, arredondo2025atoms}. However, when every atom is associated with its own node, training geometric GNNs end-to-end %\zihan{at atomistic scale} 
is memory- and runtime-intensive. For example, when learning force fields for small molecules (fewer than 100 atoms), single-GPU batch sizes are often limited to $\sim\!10^{1}$ conformations\cite{anderson2019cormorant,wang2024enhancing}. This constrains the use of geometric GNNs for analyzing biomolecular dynamics, where systems routinely involve $10^{4}$ or more atoms, and batch sizes of $10^3$ or more time-lagged pairs of conformations are needed to detect correlations spanning wide ranges of time scales~\cite{mardt_vampnets_2018,wang2021state,pengmei2025using}.

To extend GNNs to analyzing biomolecular dynamics, we recently introduced geom2vec \cite{pengmei2025using}, in which we pretrain an equivariant geometric GNN by denoising equilibrium structures of small organic molecules and use it to generate residue-level embeddings (henceforth tokens). These tokens serve as inputs to a transformer (token mixer) that is trained for the task of interest.  This architecture allows the %\zihan{atomistic} 
GNN to be trained offline with a task that demands less memory.  The tokens encode local chemical and geometric information in a way that respects molecular symmetries, and the token mixer learns task-specific nonlocal interactions among the tokens. We showed that this approach can go beyond graphs based on C$_\alpha$ atoms \cite{ghorbani2022graphvampnet,liu2023graphvampnets} to graphs based on all atoms when analyzing the folding of small proteins and, in turn, identified important side chain motions.  While geom2vec thus expands the systems and processes that can be studied with GNNs, the memory and processor time requirements of the token mixer limit applicability on a single GPU to systems of fewer than $10^2$ tokens (i.e., residues) with a minibatch size of $10^3$ time-lagged pairs of conformations.

\begin{figure}[tb]
    \centering
    \includegraphics[width=\columnwidth]{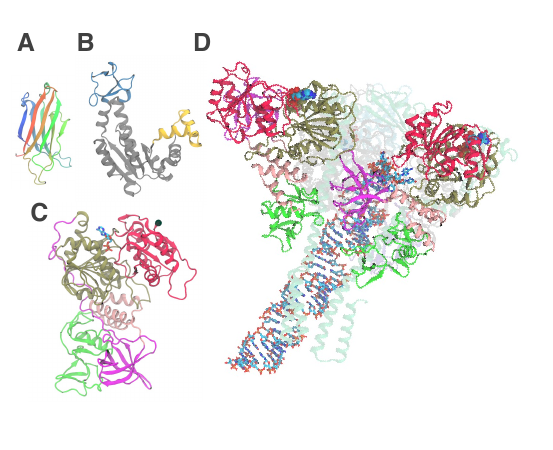}
    \caption{Systems studied in this work. \textbf{A.} The C2A domain of synaptotagmin (PDB: 2R83\cite{fuson_structure_2007}). \textbf{B.} Apo-adenylate kinase in an open conformation (PDB: 4AKE\cite{muller_adenylate_1996}). The LID, NMP, and CORE domains are shown in blue, yellow, and gray, respectively. \textbf{C.} SARS-CoV-2 helicase nsp13 bound to ADP in an open conformation\cite{chen_ensemble_2022}. The N-terminal zinc-binding domain (ZBD), stalk (S), 1B, RecA1, and RecA2 domains are shown in green, salmon, magenta, tan, and red, respectively. \textbf{D.} SARS-CoV-2 helicase replication-transcription complex in 1B-open state (PDB: 7RDX\cite{chen_ensemble_2022}). The nsp13 helicase domains are colored as in \textbf{C}. Structures are shown to scale.}
    \label{fig:systems}
\end{figure}

To address this issue, here we adapt ideas from computer vision \cite{bolya2022token} and introduce a \emph{token-merging module}.  The module consists of a graph-convolution layer that exchanges information among spatial neighbors followed by a multi-layer-perceptron (MLP) that merges windows of $w$ sequential tokens into composite (fragment-level) ones. The module thus reduces the number of pairwise interactions that the mixer must learn by a factor of $w^2$.
Beyond reducing memory and runtime requirements, this module functions as a regularizer by decreasing the tendency to fit spurious short-range token–token correlations; this improves performance (as measured by a variational loss function\cite{mardt_vampnets_2018}). The fragment abstraction also improves interpretability by shifting attention patterns to the level of several residues. 
By pairing the token-merging module with FlashAttention~\cite{dao2022flashattention}, we are able to analyze molecular dynamics simulation data for systems of up to $22\,060$ atoms (excluding solvent; Fig.~\ref{fig:systems}) on a single GPU,
%Beyond lowering compute and memory, the \tmm{} also improves model quality, as reflected by higher VAMP-2 validation scores~\cite{mardt_vampnets_2018}.
opening the door to automated interpretation of simulations of large biomolecular complexes.

\section{Methods}
\label{sec:methods}

In this section, we first review the geom2vec framework, which provides the foundation for our current work. We then describe the \tmm{} (TMM) and the use of FlashAttention~\cite{dao2022flashattention}. Subsequently, we introduce the biomolecular systems and analysis tasks employed to demonstrate our method's capabilities.

\subsection{Geom2vec framework}
\label{subsec:geom2vec_review} 

To enable analyzing biomolecular dynamics without manual feature engineering, geom2vec decouples learning local structure and nonlocal dependencies \cite{pengmei2025using}. The framework has three components:

\begin{enumerate}
    \item \textbf{Pretraining.} We pretrain an equivariant geometric GNN by denoising \cite{zaidi2022pre} perturbed structures of small organic molecules. This step leverages extensive structural data, which are much more readily available than dynamical data, to learn local structural embeddings. Moreover, because these data are not specific to the molecules of interest, we can reuse the checkpoints from our earlier studies \cite{pengmei2024geom2vec,pengmei2024pushinglimitsallatomgeometric}. The pretraining hyperparameters are listed in Table~\ref{tab:hp_pretrain}.
    \item   \textbf{Offline feature extraction.} The pretrained GNN takes Cartesian coordinates as inputs and produces atom-level embeddings that we aggregate into residue-level tokens. This step is computationally demanding but performed only once offline for a given molecule: tokens are computed and stored for reuse. These tokens capture geometric information while respecting physical symmetries (translational and rotational equivariance; permutational invariance).
    \item \textbf{Online token mixing.} A \tokenmixer{} learns nonlocal dependencies among tokens for the task of interest.
\end{enumerate}

We previously explored various architectures for the token mixer, including transformers~\cite{vaswani2017attention, pengmei2023transformers}, MLP-Mixers~\cite{tolstikhin2021mlp}, and equivariant GNNs (specifically, the geometric vector perceptron, GVP~\cite{jing2020learning}), or combinations thereof.
While an architecture that combined a transformer and a GVP performed best previously,
we subsequently found that a plain \tokenmixer{} can achieve comparable accuracy when paired with appropriate positional encodings for the GNN-derived tokens.
To simplify the architecture and avoid training an additional GVP, we therefore use only a \tokenmixer{} in this study.

A known limitation of transformers is that the computational complexity (both memory and runtime) of self-attention scales quadratically with the number of input tokens. This becomes prohibitive for large biomolecular systems that comprise thousands of residues. To address this issue, as we now describe, we introduce a TMM, which aims to reduce the effective number of tokens processed by the transformer and improve interpretability without sacrificing critical information about the system's dynamics.

\subsection{Token-Merging Module (TMM)}

\begin{figure}[htb]
    \centering
    \includegraphics[width=0.8\columnwidth]{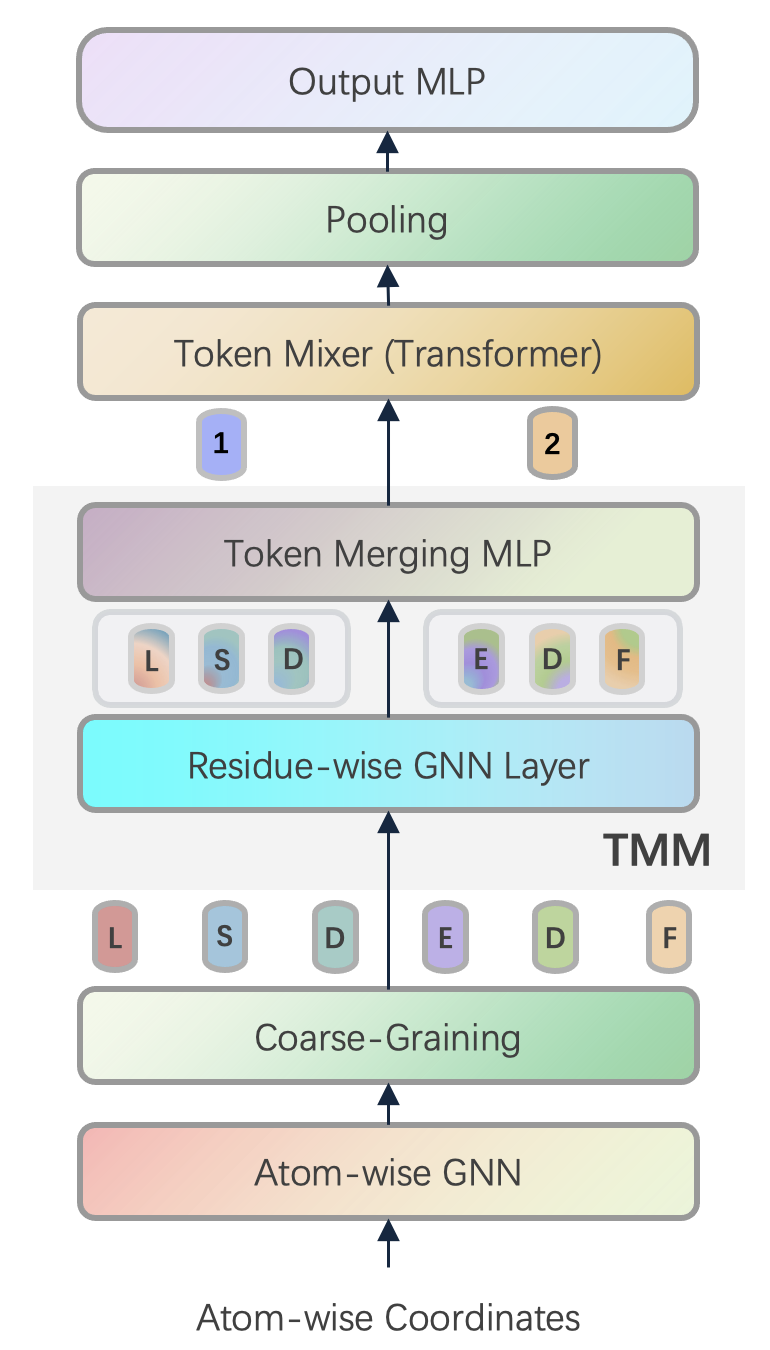}
    \caption{A schematic illustration of the geom2vec workflow with the the proposed \tmm{} (shaded). 
    The \tmm{} comprises a \fgc{} layer acting on a coarsened graph defined by residue-wise coordinates and a token-merging MLP operating on fixed windows along the sequence. A sinusoidal positional encoding is added to the reduced token set before it enters the transformer token mixer.}
    \label{fig:schema}
\end{figure}

We describe our method in terms of proteins and nucleic acids, both of which are linear polymers, but our approach can readily be adapted to other systems.  Proteins and nucleic acids are hierarchically organized.  Atoms are organized into residues, residues form secondary structures and other motifs, these assemble into tertiary structures with domains, and multiple chains come together to form quaternary structures and complexes. This structural hierarchy is generally reflected in dynamics:  local fluctuations can be distinguished from global motions. These considerations motivate our approach. Namely, we introduce a TMM that combines a graph-convolution layer with an MLP to merge windows of residue-level tokens into fragment-level tokens. This operation trades granularity for a reduction in the number of pairwise interactions that the token mixer must consider; this results not only in a large reduction in memory and runtime requirements but a small boost in performance.

The token-merging module takes residue-level tokens as inputs and outputs fragment-level tokens.  There are two parts to the token-merging module:  a graph-convolution layer that exchanges information among spatially neighboring residue-level tokens followed by an MLP that merges windows of $w$ sequential residue-level tokens into fragment-level ones.

\begin{algorithm}[H]
\small
% \setstretch{1.25}
\caption{\TMM}
\label{alg:merger}
\begin{algorithmic}[1]

\Require Residue features %of size $H$, 
$x \in \mathbb{R}^{N\times H}$, Residue positions $r\in \mathbb{R}^{N\times 3}$, window size $w$, distance cutoff $c$
\State $E \gets \operatorname{RadiusGraph}(r, c)$\Comment{Build graph connectivity}
\State $x \gets \operatorname{GraphConvLayer}(x, E)$
\State $x \gets \operatorname{Reshape}\Bigl(
    x,\ 
    \bigl(N/w, Hw\bigr) \Bigr)$ \Comment{Arrange by window}

\State $\hat{x} \gets \operatorname{MLP}(x)$ \Comment{Token merging layer}
\State \Return $\hat{x}$ \Comment{Fragment-level tokens $\hat{x}\in\mathbb{R}^{(N/w)\times H}$}

\end{algorithmic}
\end{algorithm}

\begin{enumerate}
\item{\textbf{Graph-Convolution Layer.} We construct a graph with edges between residues that are within a distance $c$ of each other;  we measure distance between C$_\alpha$ atoms for proteins and between phosphate P atoms for nucleic acids. We benchmark four graph operators:  GCNConv~\cite{kipf2016semi}, GraphConv (GC)~\cite{morris2019weisfeiler}, ResGatedGraphConv (RGGC)~\cite{bresson2017residual}, and TAGConv~\cite{du2017topology}. All follow the same aggregation/update template but differ in their neighbor weighting and normalization. We use the implementations in PyTorch Geometric~\cite{Fey/Lenssen/2019}. Full definitions and hyperparameters appear in Appendix~\ref{app:local-mp} (Eqs.~\eqref{eq:A1}–\eqref{eq:A4}).}

\item{\textbf{Token-merging MLP.}} We group the $N$ residue-level tokens into non-overlapping windows of length $w$ (except for ligands, which are assigned their own windows unless noted otherwise). Let $x\in\mathbb{R}^{N\times H}$ denote residue-level tokens after the TMM graph-convolution layer. We reshape to $(N/w,wH)$ and apply an MLP that maps $wH$ to $H$ to yield one \fragtoken{} per window:
\begin{equation}
\mathbf{x}_{\mathrm{merged}} \;=\; 
\operatorname{MLP}\!\Bigl(\,
  \operatorname{Concat}\bigl[\mathbf{x}_i,\ldots,\mathbf{x}_{i+w-1}\bigr]
\Bigr).
\end{equation}
If $N$ is not divisible by $w$, we pad the final window to length $w$ with zeros. Each \fragtoken{} has dimensionality $H$ (cf.\ Algorithm~\ref{alg:merger}). We add fragment-level positional encodings after merging and then pass the \fragtoken s to the \Transformer.
\end{enumerate}

Together, the \fgc{} layer and the \wmlp{} form the TMM, a plug-in component trained end-to-end with the \Transformer{} to shorten the input sequence ($N \!\to\! N/w$) and reduce the number of self-attention pairs.

\subsection{Reducing memory cost with efficient self-attention}
\label{subsec:flash_attention}

The \tokenmixer{} determines dependencies among the fragment-level tokens output by the \tmm{} using a self-attention mechanism \cite{vaswani2017attention}
\begin{align}
\operatorname{Attention}\bigl(\mathbf{Q},\mathbf{K},\mathbf{V}\bigr)
\;=\;
\operatorname{softmax}\!\bigl(\mathbf{Q}\mathbf{K}^\tran / \sqrt{d}\bigr)\,\mathbf{V},
\end{align}
where $\mathbf{Q}$, $\mathbf{K}$, and $\mathbf{V}$ are the standard query, key, and value matrices, respectively. Although token merging reduces the sequence length from $N$ to $N/w$, naive self-attention would 
still require an $(N/w)\times(N/w)$ attention map. For large biomolecules, this quickly becomes computationally
expensive.
We address this issue with FlashAttention\cite{dao2022flashattention}, which computes 
the exact softmax-based attention blockwise in high-speed on-chip GPU memory so as to avoid storing the entire 
$\mathbf{Q}\mathbf{K}^\tran$ matrix \cite{dao2022flashattention}. FlashAttention reduces the extra memory overhead from $\mathcal{O}(N^2d)$ to $\mathcal{O}(Nd)$, which can result in significant savings for long sequences.

\subsection{Analyzing dynamics}

We evaluate the performance of the \tmm{} with different graph operators using the variational approach for Markov processes (VAMP) for identifying slow modes of motion \cite{mardt_vampnets_2018, chen_nonlinear_2019,wu_variational_2020, lorpaiboon_integrated_2020} because its variational nature makes it easy to compare results. We then demonstrate the use of the best performing architectures with the state predictive information bottleneck (SPIB) for identifying metastable states \cite{wang2021state}. However, we stress that our approach is general and can be applied to other objectives, such as computing kinetic statistics \cite{strahan_predicting_2023,strahan_inexact_2023,chen_discovering_2023}. For completeness, we briefly review VAMP and SPIB.

\subsubsection{VAMP}

Given a Markovian dynamics $\mathbf{X}_t$ (as generated by molecular dynamics), one can define the correlation functions
\begin{align}
    C_{00} &= \E_{\mathbf{X}_0 \sim \nu}[\chi_0(\mathbf{h}(\mathbf{X}_0)) \chi_0^\tran(\mathbf{h}(\mathbf{X}_0))]  \\
    C_{0\tau} &= \E_{\mathbf{X}_0 \sim \nu}[\chi_0(\mathbf{h}(\mathbf{X}_0))\chi_\tau^\tran(\mathbf{h}(\mathbf{X}_\tau))]  \\
    C_{\tau\tau} &=\E_{\mathbf{X}_0 \sim \nu}[\chi_\tau(\mathbf{h}(\mathbf{X}_\tau)) \chi_\tau^\tran(\mathbf{h}(\mathbf{X}_\tau))],
\end{align}
where $\chi_0$ and $\chi_\tau$ are vectors of functions, and the expectation is over trajectories initialized from an arbitrary distribution $\nu$. In the original implementation of VAMPnets\cite{mardt_vampnets_2018}, the arguments to $\chi_0$ and $\chi_\tau$ were either the Cartesian coordinates of aligned structures or distances between non-hydrogen atoms; in the present study, $\mathbf{h}(\mathbf X)$ is the output of the token mixer.  That is, we learn the arguments to $\chi_0$ and $\chi_\tau$. VAMP seeks the $\chi_0$ and $\chi_\tau$ that maximize the correlation functions so that they approximate the slowest decorrelating modes. We specifically maximize the VAMP-2 score:
\begin{equation} \label{eq:vamp2}
    \operatorname{VAMP-2} = \left\lVert C_{00}^{-1/2} C_{0\tau} C_{\tau\tau}^{-1/2} \right\rVert_{\rm F}^2,
\end{equation}
where the subscript $\rm F$ denotes the Frobenius norm.

\subsubsection{SPIB}

In the information bottleneck (IB) framework, an encoder-decoder setup is used to learn a low-dimensional (latent) representation $\mathbf{z}$ that minimizes the information from high-dimensional inputs while maximizing the information about targets. In the SPIB extension of IB, the inputs are the molecular features $\mathbf{h}(\mathbf{X}_t)$ at time $t$, and the targets are the state labels $s_{t+\tau}$ at time $t+\tau$.
To learn the latent representation $\mathbf{z}$, SPIB maximizes the loss function
%\begin{widetext}
\begin{multline}
    \label{eq:spib}
    \mathcal{L}_\text{SPIB} = \E_{\mathbf{X}_0, \mathbf{z}}
    \Biggl[
        \ln q_\theta(s_{\tau}|\mathbf{z})
        - \beta \ln \frac{p_\theta(\mathbf{z}|\mathbf{h}(\mathbf{X}_0))}{r_\theta(\mathbf{z})}
    \Biggr].
\end{multline}
where the expectation is over $\mathbf{X}_0 \sim \nu$, where $\nu$ is an arbitrary sampling distribution, and $\mathbf{z} \sim p_\theta(\mathbf{z}|\mathbf{h}(\mathbf{X}_0))$.
%\end{widetext}
The encoder generates the latent representation $\mathbf{z}$ from the input with probability $p_\theta(\mathbf{z}|\mathbf{h}(\mathbf{X})) = \mathcal{N}(\mathbf{z};\mu_\theta(\mathbf{h}(\mathbf{X})),\Sigma_\theta(\mathbf{h}(\mathbf{X})))$, which is a multivariate normal distribution with learned mean $\mu_\theta(\mathbf{h}(\mathbf{X}))$ and learned covariance $\Sigma_\theta(\mathbf{h}(\mathbf{X}))$. The decoder takes the latent representation $\mathbf{z}$ and returns the probability $q_\theta(s|\mathbf{z})$ of each state label $s$; $q_\theta(s|\mathbf{z})$ is represented by a neural network with output dimension equal to the number of possible state labels.
The quantity $r_\theta(\mathbf{z})$ is a prior.
The state labels are updated during training as follows:
\begin{equation}
    \label{eq:spib_refine}
    s_\tau = \operatorname{argmax}_s q_\theta(s|\mu_\theta(\mathbf{h}(X_\tau))).
\end{equation}
We follow \citet{wang2021state} and our earlier work\cite{pengmei2025using} and use a variational mixture of posteriors for the prior \cite{tomczak2018vae}.
In this work, we prepared the initial state labels by performing \textit{k}-means clustering on the coordinates learned from VAMPnets %based on SubFormer-Merger (SF-Merger) \cite{pengmei2023transformers,pengmei2025using} 
with $k=100$ clusters.

\subsection{Datasets}
\label{sec:systems}
We examine the performance of the proposed method by analyzing data from molecular dynamics simulations of four systems:   synaptotagmin C2A (128 residues), adenylate kinase (214 residues), the SARS-CoV-2 nsp13 helicase (592 residues), and the SARS-CoV-2 replication-transcription complex (2645 residues).  Because we mainly present results for adenylate kinase (ADK) and nsp13, we discuss them in more detail here.

\subsubsection{Adenylate kinase}\label{sec:sys_adk}
Adenylate kinase (ADK) catalyzes interconversion of adenosine mono-, di-, and triphosphates (AMP, ADP, and ATP). Its structure consists of three domains: a core domain (CORE, residues 1--6, 14--29, 68--119, 161--214), an ATP-binding domain (LID, residues 120--160), and an AMP-binding domain (NMP, residues 30--67); residues 7--13 are known as the P-loop. Crystal structures with and without inhibitors  suggest that the LID and NMP domains adopt open and closed conformations\cite{muller_structure_1992, muller_adenylate_1996}. 
These dynamics have been extensively studied through simulations\cite{beckstein_zipping_2009, seyler_path_2015, zheng_multiple_2018a} and experiments\cite{stiller_probing_2019, stiller_structure_2022}. 

Here, we analyze approximately 54 $\mu$s of unbiased simulations (sampled as 70 trajectories ranging in length from 350 to 1320 ns) of the apo-enzyme from \citet{zheng_multiple_2018a}, who constructed an MSM of the conformational changes in ADK. We use their simulations performed with the AMBER ff99SBnmr1 force field\cite{li2011iterative}. % in explicit TIP3P solvent at a temperature of 310~K. 
Configurations were saved every 20~ps.

\subsubsection{SARS-CoV-2 nsp13}\label{sec:sys_nsp}

The SARS-CoV-2 viral genome is replicated and transcribed by a complex of nonstructural protein (nsp) 7, two copies of nsp8, and nsp12. This complex together with RNA forms a replication-transcription complex (RTC) that associates with two copies of nsp13, a helicase that unwinds double-stranded DNA and RNA \cite{chen_ensemble_2022, chen_structural_2020, yan_architecture_2020, yan_cryoem_2021}.
Ensemble cryo-EM analysis of the nsp13\textsubscript{2}-RTC complex shows multiple conformations of nsp13 interacting with RNA. In its engaged (apo) state, nsp13 traps a strand of RNA between its RecA1, RecA2, and 1B domains (Figure~\ref{fig:systems}C), while in its open state (1B-open\cite{chen_ensemble_2022}) the 1B domain rotates around the stalk away from the RecA1 and RecA2 domains to create an open binding channel. In the full nsp13\textsubscript{2}-RTC complex, the open 1B domain of one copy of nsp13 is prevented from fully closing by the other copy.

Here, we analyze five trajectories of free nsp13 bound to ADP-$\ce{Mg^2+}$ (nsp13-ADP) initialized from the 1B-open conformation  \cite{chen_ensemble_2022}. Each of the trajectories is nearly 25 $\mu$s in length, with configurations saved every 1.2 ns. Three of the trajectories contain a spontaneous transition to the 1B-closed conformation. %Proteins, RNAs, and ions were parameterized with the DES-Amber SF1.0 force field, and water was represented by the TIP4P-D model. 
%Simulations were performed at 310 K and .

\subsubsection{Training-validation split}\label{sec:split}

As previously discussed \cite{pengmei2024beyond, pengmei2025using}, due to the strong correlation between successive structures sampled by molecular dynamics simulations, a random split of the data into training and validation sets allows networks to achieve high validation scores even when they have memorized the training data rather than learned useful abstractions from it; the models then perform poorly on independent data. Consequently, for ADK, we randomly draw 20\% of the trajectories as the hold-out validation set. For nsp13-ADP, we temporally split each of five independent trajectories into two fragments and then randomly draw 20\% of all 10 fragments as the hold-out validation set.

\section{Results}

\subsection{VAMP-based comparison of architectures}

\begin{figure*}[htbp]
    \centering
    \includegraphics[width=0.8\textwidth]{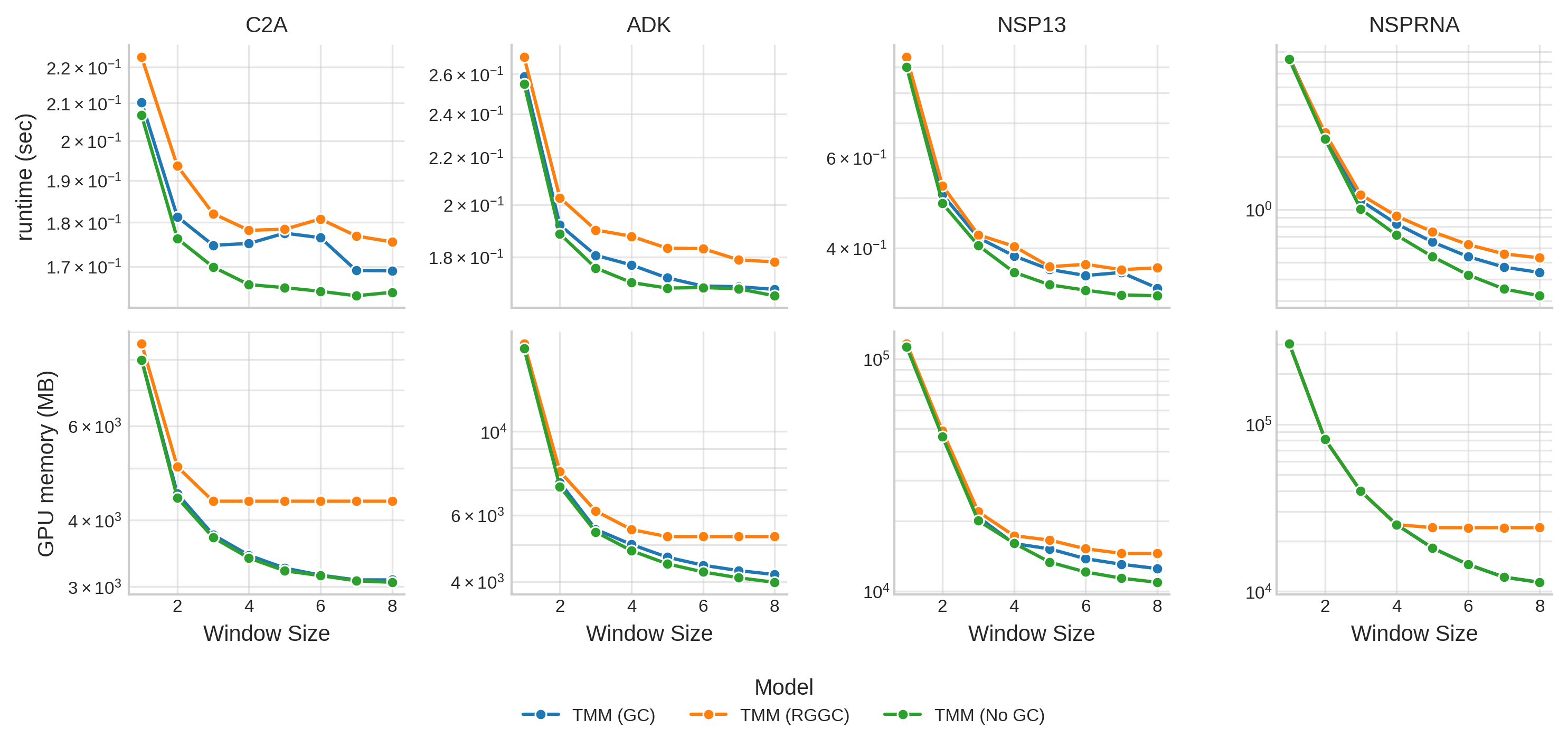}
     % \caption{GPU performance of SF-Merger model variants across varying window sizes. The top row displays forward pass runtime, and the bottom row shows peak GPU memory usage, both normalized to a throughput of 1000 samples. Systems evaluated are: synaptotagmin C2A (128 fragments), ADK (216 fragments), nsp13-ADP (592 fragments), and nsp13-RNA (2645 fragments). All metrics were averaged over multiple runs following a warm-up. Measurements for C2A, ADK, and nsp13-ADP were performed on an NVIDIA A40 GPU (48 GB memory), using batch sizes that allowed for this normalization (e.g., batch size 1000 for C2A/ADK, 50 for nsp13-ADP) with a fixed feature dimension of 64 and the same configuration. The nsp13-RNA system was benchmarked on an NVIDIA A100 GPU (40 GB memory) with a batch size of 50, with results similarly normalized. The plots illustrate a shift from quadratic-like scaling (small window sizes) towards more computationally efficient, sub-quadratic scaling as window size increases, demonstrating the effectiveness of token merging in reducing self-attention costs.}
     \caption{Runtime (top) and peak GPU memory (bottom) for TMM as functions of window size $w$ for synaptotagmin C2A (128 residues), ADK (214 residues), nsp13-ADP (592 residues), and nsp13-RNA (2645 residues).  Numbers reported are for a minibatch size of 1000 samples; for nsp13-ADP and nsp13-RNA, we use a minibatch size of 50 in practice and then scale the measured values to those for a minibatch size of 1000.}

    \label{fig:profile}
\end{figure*}

\begin{figure*}[htbp]
    \centering
    \includegraphics[width=0.75\textwidth]{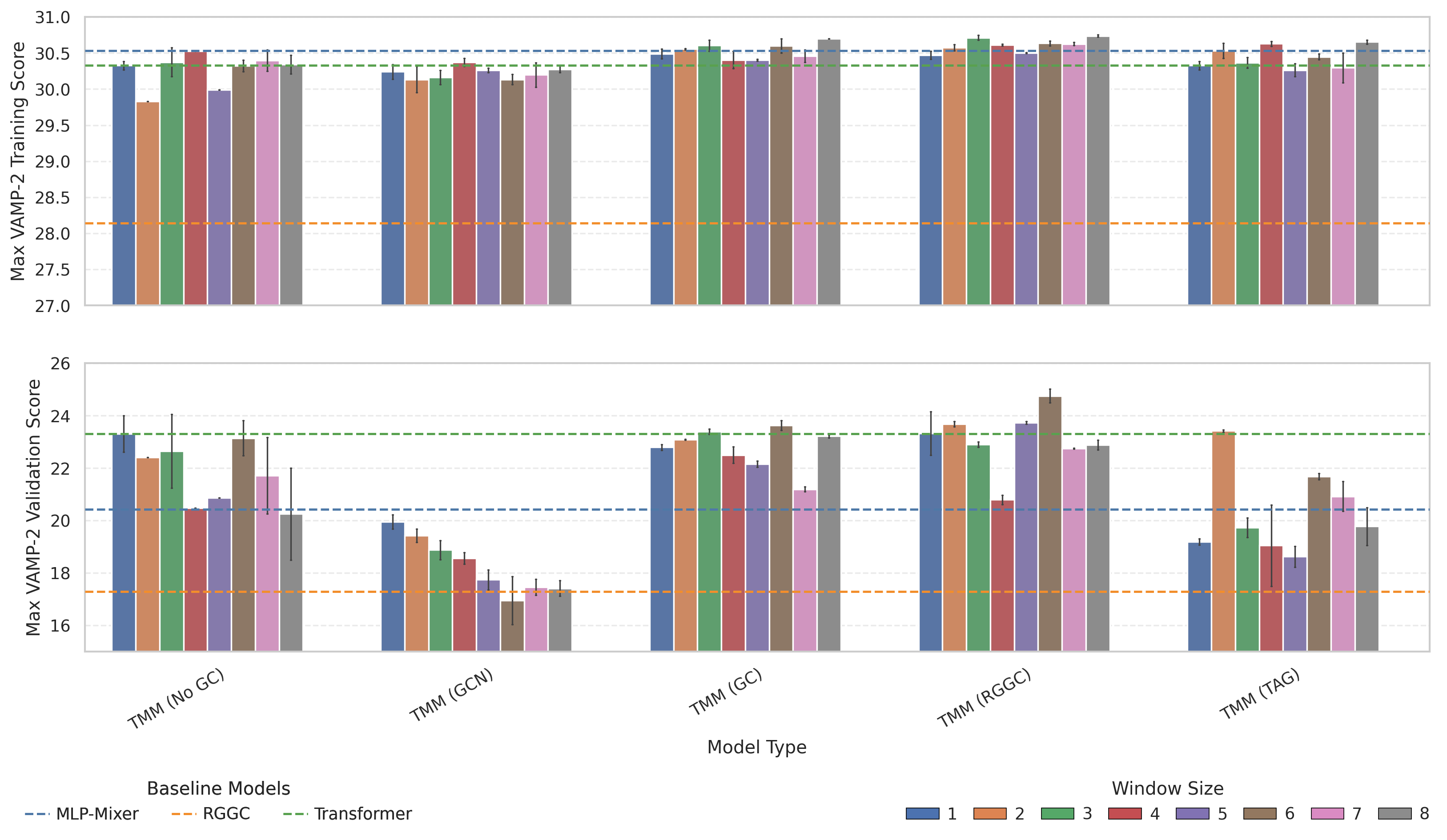}
    \caption{Maximum training (top) and validation (bottom) VAMP-2 scores for ADK.  Each group of bars corresponds to a different graph operator for the TMM graph-convolution layer (see Appendix \ref{app:local-mp}). A window size of 1 is equivalent to no token merging. The training and validation data are held fixed across repeated runs and models. ``RGGC'' is a GNN baseline consisting of 4 RGGC layers with mean pooling (no token mixer). %The global token mixer is a \Transformer{} with TMM.   
    Error bars show standard deviations over three runs.}
    \label{fig:val_score_adk}
\end{figure*}

\begin{figure*}[htbp]
    \centering
    \includegraphics[width=0.7\linewidth]{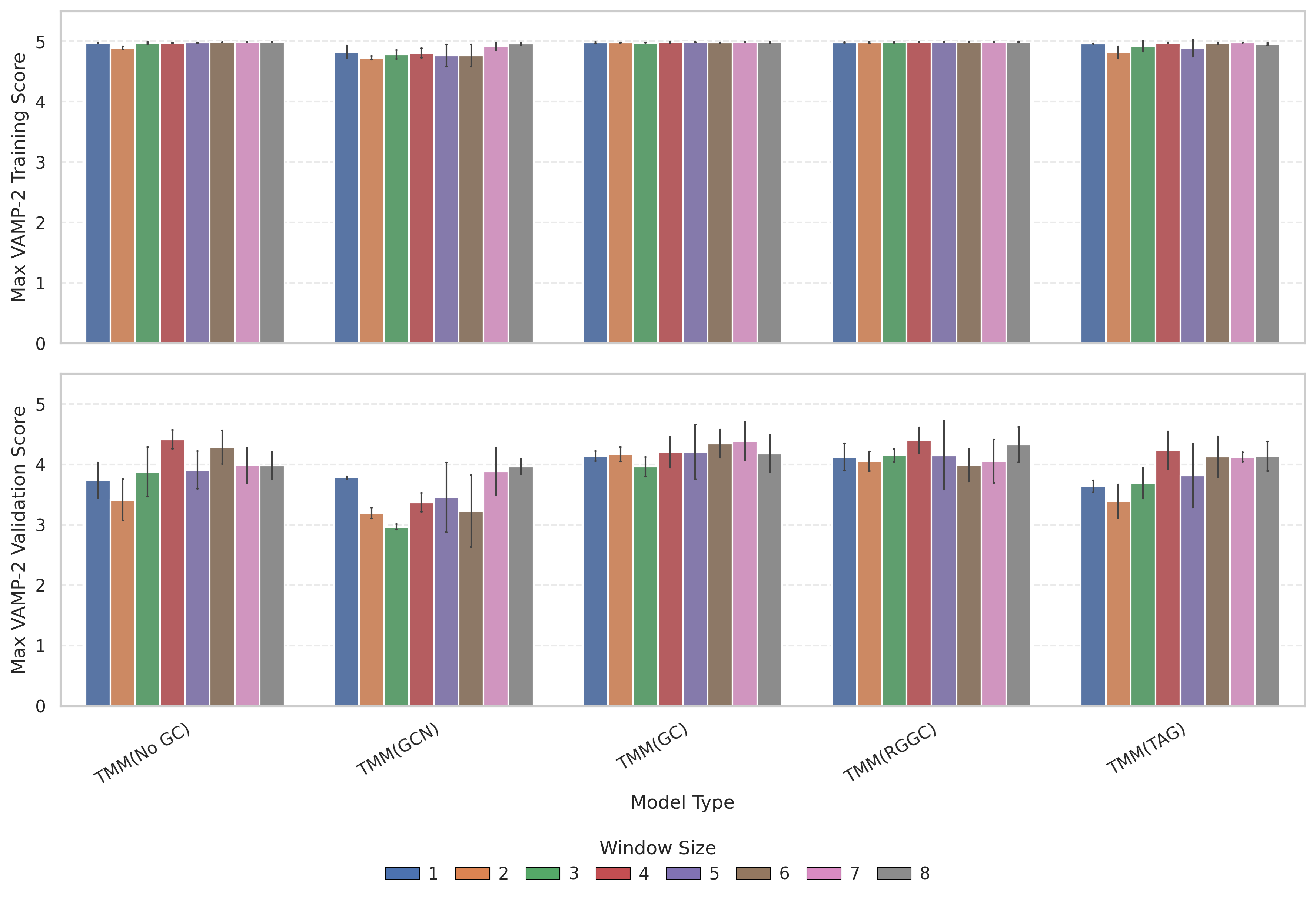}
    \caption{Maximum training (top) and validation (bottom) VAMP-2 scores for nsp13-ADP. One validation step is taken every 40 training steps. Each group of bars corresponds to a different graph operator for the TMM graph-convolution layer (see Appendix \ref{app:local-mp}). A window size of 1 is equivalent to no token merging. Error bars show standard deviation over three runs. }
    \label{fig:curves_nsp13}
\end{figure*}

In this section, we systematically explore how architectural choices affect efficiency and performance.  
To this end, 
the memory and runtime requirements for the four systems are shown in Figure \ref{fig:profile}.  As expected, these requirements increase with system size and decrease with window size.  In general, the decreases with window size are larger for smaller window sizes, and, for the RGGC architecture in particular, the restricted gating results in a floor to the memory improvements.  

We quantify performance by the VAMP-2 score, which measures the quality of approximations of the slowest decorrelating modes, as described in Section \ref{sec:methods}. A higher VAMP-2 score is better.
Figures \ref{fig:val_score_adk}, \ref{fig:curves_nsp13}, and \ref{fig:scores_c2a} show the maximum training and validation VAMP-2 scores for ADK, nsp13, and synaptotagmin C2A for various graph operators for the TMM graph-convolution layer, averaged over multiple runs, with colors representing different window sizes. 
For ADK and synaptotagmin C2A, we additionally show two baseline models.
One comprises four RGGC layers as the token mixer; all the models outperform this baseline, indicating that a long-range token mixer is essential for capturing the features of the dynamics.
The other is a model that lacks a TMM and has an MLP-Mixer \cite{tolstikhin2021mlp} as the token mixer; we choose this model as a baseline because it scales subquadratically but has a global receptive field.

The differences between models are mainly in the validation scores, indicating that all the models are sufficiently expressive to (over)fit the training data but vary in their ability to generalize.
%consistent with the discussion of the train-validation split above (Section~\ref{sec:split}). 
Specifically, the simpler GC and RGGC update mechanisms outperform the more complex GCN (with self-loops and normalization) and TAGConv (with multi-hop contributions) update mechanisms.  The stronger graph-inductive biases of GCN and TAGConv may reduce their generalization capabilities. 
Importantly, token merging does not diminish the ability to learn the relevant kinetics. In fact, it appears to increase the VAMP-2 scores beyond those that can be achieved without it (window size of 1). A possible explanation is that token merging provides a form of regularization by reducing the lengths of sequences and forcing the transformer to focus on nonlocal dependencies.

\subsection{SPIB}

Having established the efficacy of our architecture through VAMP-2 metrics, we next demonstrate its applicability to downstream analysis tasks. Specifically, we employ SPIB on the ADK and nsp13-ADP datasets, using the hyperparameters that yielded the highest VAMP-2 scores.  Given the states from SPIB, we construct MSMs by counting transitions between the states. We do not symmetrize the transition matrices to enforce microscopic reversibility.

\subsubsection{ADK}

\begin{figure*}[htbp]
    \centering
    \includegraphics{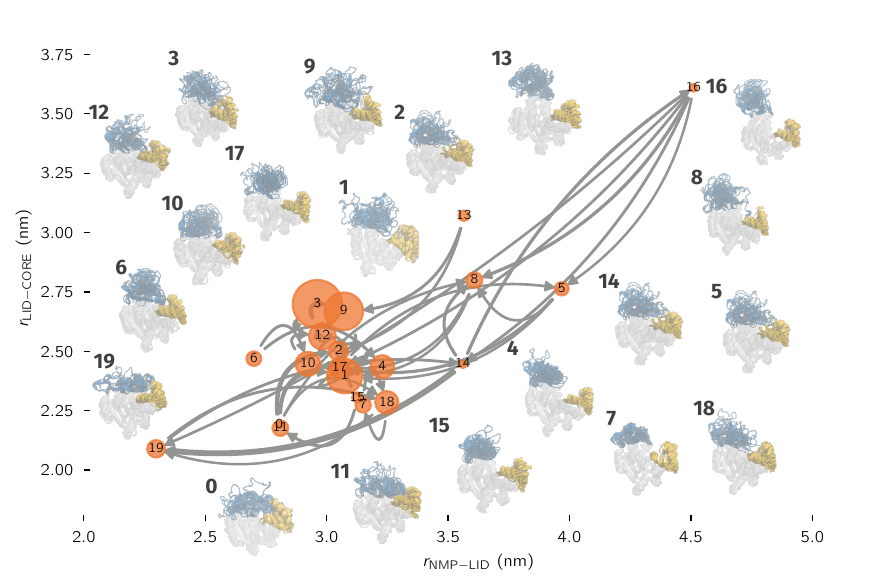}
    \caption{MSM of SPIB-learned states for ADK estimated with a lag time of 10~ns. Each labeled node represents a metastable state, and the size of each node is proportional to its population.  The edges indicate the highest-probability transitions between states, with thicker edges denoting a larger number of transitions per unit time.  For each state, we show 15 random structures that are aligned by their CORE domains. The states are plotted according to the average center of mass distances between the NMP and LID domains or between the LID and CORE domains.}
    \label{fig:spib_network_adk}
\end{figure*}

\begin{figure*}[htbp]
    \centering
    \includegraphics[width=0.8\textwidth]{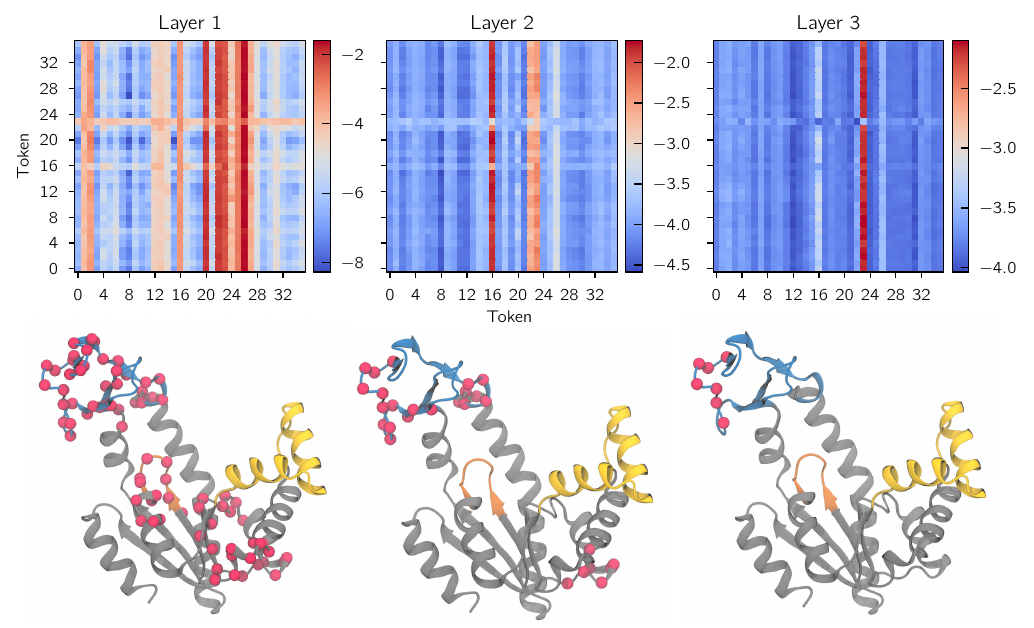}
    \caption{ (Top) Maps of log-attention weights from each layer of a transformer (with TMM) averaged across all SPIB conformational states and across attention heads. Each index is a \fragtoken{} produced with $w=6$ (i.e., 6 residues per fragment). (Bottom) ADK (PDB: 4AKE\cite{muller_adenylate_1996}) open structure with LID domain (fragments 120--160, blue), NMP domain (fragments 30--67, yellow), P-loop (fragments 7--13, orange), and CORE domain (gray) colored. Fragment positions with high attention weights are indicated with red spheres.  }
    \label{fig:adk_attn}
\end{figure*}

For the SPIB analysis of the ADK dataset, we use a model with the RGGC update mechanism and a window size of $w=6$. The training time for this model on an Nvidia A40 GPU was 28.9 min, compared with 84.4 min for a model without TMM.  The acceleration is less than $w^2=36$ fold because ADK is only 214 residues, so the overhead is still a significant part of the runtime.  

The analysis yields 20 metastable states, which are plotted  in Figure~\ref{fig:spib_network_adk} according to the distances between the centers of mass of the NMP and LID domains ($r_{\mathrm{NMP-LID}}$) and between centers of mass of the LID and CORE domains ($r_{\mathrm{LID-CORE}}$), following \citet{zheng_multiple_2018a}. Consistent with their MSM, the states learned by SPIB can be grouped into three broad conformational classes: (i) open-like (upper right), (ii) intermediate (middle), and (iii) closed-like (lower left).
The open-like states resemble the fully open crystal structure, featuring large separations of the LID and NMP domains and minimal contact between the LID and CORE domains, with the LID domain displaying greater heterogeneity. The intermediate states typically have a partially closed LID domain (to varying degrees) while the NMP domain remains relatively open (Figure~\ref{fig:spib_network_adk_theta}). The closed-like states are more compact, with extensive contact between the LID and CORE domains.

The transition network between the open-like and closed-like states, indicated by the edges in Figure~\ref{fig:spib_network_adk} and \ref{fig:spib_network_adk_theta}, is also consistent with the analysis of Zheng and Cui \cite{zheng_multiple_2018a}. Open-like metastable states (especially 5, 8, and 16) rapidly interconvert while also feeding into several intermediate states (1, 9, 14) with partially closed LID domains, which form a central ``hub'' of the conformational network.  The hub is in turn connected to the closed-like states  through several states (2, 10, and 15), indicating that the open-close transition of ADK does not follow a single pathway.

To characterize the internal representations of the model, we examine the averaged attention weights for the \tokenmixer{}, which has three layers.  Key fragments appear as red rows/columns in the heatmaps (Figure \ref{fig:adk_attn}). In layer 1, the network primarily attends to three major elements: the LID domain (tokens 20--26), the P‐loop (tokens 7–13), and the parts of the CORE (tokens 1, 2, and 16). The residues in the CORE domain, as well as those at the LID-CORE hinge (tokens 20 and 26, corresponding to residues 121--126 and 157--162) correspond to contacts identified by Zheng and Cui as having high mutual information with MSM transition times (compare with Figure S19 in ref.~\citenum{zheng_multiple_2018a}), suggesting that the model is able to identify physically meaningful information automatically. 
%The attention weights for fragments in the NMP domain are small, consistent with the relatively smaller motions of the NMP domain compared to the LID in the MSM analysis. 
By layer 2, the attention narrows to specific elements of the LID and CORE domains, suggesting that the model refines its notion of which tokens mediate inter‐domain contacts. Finally, in layer 3, attention becomes even more localized to the LID loop, which is flexible during opening events. 

% By contrast, no interpretable pattern emerges without the TMM (Figure \ref{fig:spib_plain_transformer_attn}). 

\begin{figure}[h]
    \centering
    \includegraphics[width=0.8\linewidth]{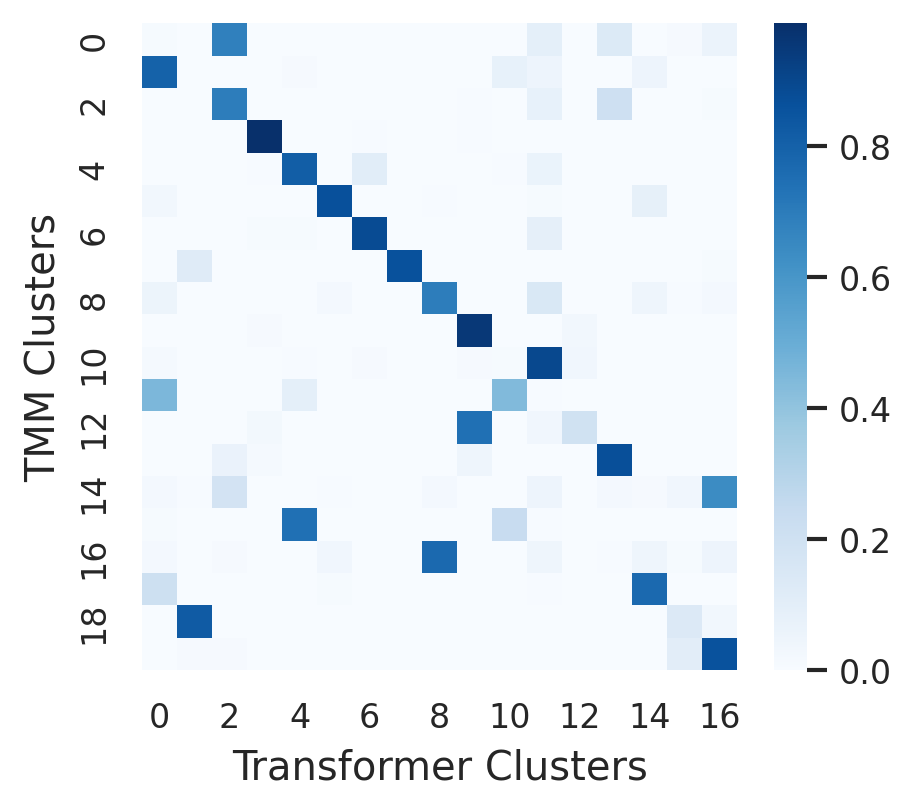}
    \caption{
    Overlap of ADK SPIB metastable states obtained with and without TMM (confusion matrix).}
    \label{fig:confusion}
\end{figure}

It is worth comparing these attention maps with ones obtained from a model without TMM (Figure \ref{fig:spib_plain_transformer_attn}).  The maps with and without TMM are generally consistent at each layer, but the attention is spread more evenly over the residues in the latter case, which makes it harder to interpret and presumably accounts for the model's weaker performance.  Examining the outputs of the models, the metastable states identified by the two models are generally consistent, but the model without TMM merges some of the states from the model with TMM (Figure \ref{fig:confusion}), consistent with a decrease in ability to resolve details.  Put together, these results suggest that TMM increases the information that can be gained in addition to providing computational benefits.

\subsubsection{Nsp13-ADP}

\begin{figure}[htb]
    \centering
    \includegraphics[width=\columnwidth]{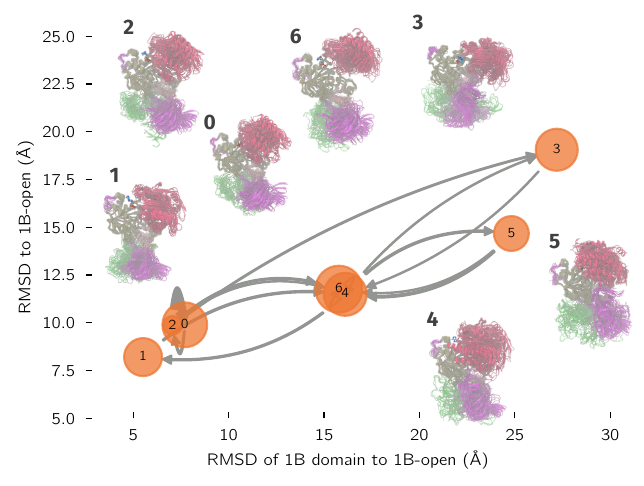}
    \caption{MSM of SPIB-learned states for nsp13-ADP estimated with a lag time of 30~ns. For each state, we show 15 random structures that are aligned by their RecA1 domains. The states are plotted according to their average non-hydrogen atom RMSD of the 1B domain (tokens 150--228) and full protein with respect to the 1B-open cryo-EM structure (PDB: 7RDX\cite{chen_ensemble_2022}), after aligning to the RecA1 domain (tokens 240--440).  Colors indicate domains: 1B (magenta), RecA1 (tan), RecA2 (red), and ZBD (green).}
    \label{fig:spib_network_nsp}
\end{figure}

For the SPIB analysis of the nsp13-ADP trajectories, we use a model with the GC update mechanism and a window size of $w=6$. The analysis yields 7 states, which we characterize by their 1B domain and total RMSD to the 1B-open cryo-EM structure, after aligning to the RecA1 domain (Figure~\ref{fig:spib_network_nsp})\cite{chen_ensemble_2022}. Based on their 1B domain conformations, the states learned by SPIB can be grouped into three classes:  open (states 0, 1, 2), intermediate (states 4 and 6), and closed (states 3 and 5). In the open states, the 1B domain (magenta) lies away from the RecA1 and RecA2 domains (tan and red), preventing nsp13 from engaging an RNA molecule in the replication-transcription complex. These states differ in the orientation of the ZBD (green) and RecA2 (red) domains. The ZBD domain appears fairly flexible, whereas the RecA2 domain can rotate toward or away from RecA1, sampling structures that are either deactivated or primed to engage with RNA (similar to the nsp13-apo or nsp13-engaged states previously seen in cryo-EM\cite{chen_ensemble_2022}). Interestingly, the rate of interconversion between states 0/2 and state 1 is very small despite their similar 1B domain conformations. The scarcity of transitions observed in the five trajectories likely accounts for the missing connectivity.

Intermediate states 4 and 6 both display moderate degrees of 1B rotation, but differ in the orientation of their RecA2 domains similar to the differences between states 0 and 2, with state 6 displaying a conformation consistent with nsp13-apo and state 4 displaying a conformation consistent with nsp13-engaged. These states primarily serve as the hub of the nsp13-ADP network, interconverting between both open and closed states. Finally, the closed states 3 and 5 show different orientations of the 1B, despite being both rotated toward the RecA domains.

\section{Discussion}
\label{sec:computational_requirements}

Molecular dynamics simulations generate high-dimensional data that are difficult to interpret. While deep-learning frameworks based on geometric GNNs hold promise for automatically identifying important features, their computational demands previously restricted their applicability to small systems. Here, we introduce a token merging scheme that compresses local structural information in a way that does not compromise and can even improve overall performance of methods for analyzing dynamics. This scheme, together with FlashAttention \cite{dao2022flashattention}, enables us to treat large biomolecular complexes.  We specifically tested our approach for VAMPnets and SPIB, but it is general and can be applied to other deep learning tasks for molecular simulations.

The geom2vec framework that we introduced previously separates training the GNN from the specific data and task of interest.  This has the advantage that we can use both data and a batch size that are computationally tractable for pretraining.  Given the trained GNN, the data of interest can be featurized offline. This leaves training the token mixer as the main online computational cost. In the case that the token mixer is a transformer, the computational cost scales quadratically with the number of input tokens, which in our original implementation \cite{pengmei2024geom2vec} was the number of residues.  The merger module does not change the fundamental scaling of the token mixer, but by grouping windows of tokens together it reduces the number of inputs to the token mixer by a factor of the window size, $w$.  In the case of a transformer token mixer, this reduces the computational cost by a factor of $1/w^2$.  In the present case, we grouped tokens sequentially, but other proximity-based schemes can be imagined. TMM could also be stacked for further savings.  These possibilities are worth investigating in the future.

It is worth contrasting our approach with ones that have been introduced specifically to circumvent or mitigate the ``shortsightedness'' of GNNs. 
Generally they introduce a form of attention.  
For example, Graphormer \cite{ying2021transformers,shi2022benchmarking, zheng2024predicting}, a transformer, biases the self-attention calculation with geometric information for long-range atom pairs. Though intuitive, this approach has limited ability to scale to large systems due to its incompatibility with FlashAttention kernels. Ewald/grid-based methods \cite{kosmala2023ewald, wang2024neural} and virtual nodes \cite{caruso2025extending} with message-passing inject global information without quadratic complexity but are specialized architectures. Here, we adapt ideas from hierarchical models that combine short- and long-range interactions \cite{Fey/etal/2020, pengmei2023transformers,li2023long, abramson2024accurate} to capture long-range dependencies with a standard transformer. This architecture has the advantages that efficient implementations are available and its attention maps are interpretable. In particular, as we showed, the attention maps can reveal protein structural organization, consistent with previous observations for protein sequence data\cite{rives2019biological,lin2023evolutionary}.

Another approach that seeks to facilitate analyzing large systems is VAMPnets combined with independent Markov decomposition (iVAMPnets) \cite{mardt2022deep}.  
This approach aims to represent an overall dynamics in terms of Markov models for subsystems that behave approximately independently.  The decomposition aids human understanding and appears to enable more data-efficient learning, though it remains unclear how to account for couplings between the subsystems.  The geom2vec with TMM that we introduce here does not make any assumption of independence and, given that the learned representations appear to be physically interpretable, they can potentially reveal couplings.  In this sense, the iVAMPnets and geom2vec with token merging approaches are complementary.  They could either be applied independently and compared, or geom2vec with token merging could serve as an expressive architecture for iVAMPnets.  

The advances described here pave the way for studying previously intractable systems, such as viral replication complexes or multi-domain enzymes, while offering a modular foundation for integration with enhanced sampling or multiscale modeling. The method’s open-source implementation ensures accessibility, inviting broader adoption in mechanistic studies and therapeutic design.

\section*{Acknowledgments}
We thank D. E. Shaw Research and Qiang Cui for making their molecular dynamics trajectories available to us.  This work was supported by National Institutes of Health award R35 GM136381.
 This work was completed with computational resources administered by the University of Chicago Research Computing Center, including Beagle-3, a shared GPU cluster for biomolecular sciences supported by the NIH under the High-End Instrumentation (HEI) grant program award 1S10OD028655-0.\\

\section*{Data availability}
Data sharing is not applicable to this article as no new data were created or analyzed in this study.
Code for our implementation and examples are available at \url{https://github.com/dinner-group/geom2vec}.

\section*{Supplementary Material}
Hyperparameter choices, additional results on synaptotagmin C2A, and supplementary figures can be found in the Supplementary Material.

\section*{References}
%\bibliography{ref}

\begin{thebibliography}{75}%
\makeatletter
\providecommand \@ifxundefined [1]{%
 \@ifx{#1\undefined}
}%
\providecommand \@ifnum [1]{%
 \ifnum #1\expandafter \@firstoftwo
 \else \expandafter \@secondoftwo
 \fi
}%
\providecommand \@ifx [1]{%
 \ifx #1\expandafter \@firstoftwo
 \else \expandafter \@secondoftwo
 \fi
}%
\providecommand \natexlab [1]{#1}%
\providecommand \enquote  [1]{``#1''}%
\providecommand \bibnamefont  [1]{#1}%
\providecommand \bibfnamefont [1]{#1}%
\providecommand \citenamefont [1]{#1}%
\providecommand \href@noop [0]{\@secondoftwo}%
\providecommand \href [0]{\begingroup \@sanitize@url \@href}%
\providecommand \@href[1]{\@@startlink{#1}\@@href}%
\providecommand \@@href[1]{\endgroup#1\@@endlink}%
\providecommand \@sanitize@url [0]{\catcode `\\12\catcode `\$12\catcode
  `\&12\catcode `\#12\catcode `\^12\catcode `\_12\catcode `\%12\relax}%
\providecommand \@@startlink[1]{}%
\providecommand \@@endlink[0]{}%
\providecommand \url  [0]{\begingroup\@sanitize@url \@url }%
\providecommand \@url [1]{\endgroup\@href {#1}{\urlprefix }}%
\providecommand \urlprefix  [0]{URL }%
\providecommand \Eprint [0]{\href }%
\providecommand \doibase [0]{https://doi.org/}%
\providecommand \selectlanguage [0]{\@gobble}%
\providecommand \bibinfo  [0]{\@secondoftwo}%
\providecommand \bibfield  [0]{\@secondoftwo}%
\providecommand \translation [1]{[#1]}%
\providecommand \BibitemOpen [0]{}%
\providecommand \bibitemStop [0]{}%
\providecommand \bibitemNoStop [0]{.\EOS\space}%
\providecommand \EOS [0]{\spacefactor3000\relax}%
\providecommand \BibitemShut  [1]{\csname bibitem#1\endcsname}%
\let\auto@bib@innerbib\@empty
%</preamble>
\bibitem [{\citenamefont {Singer}\ and\ \citenamefont
  {Sigworth}(2020)}]{singer_computational_2020}%
  \BibitemOpen
  \bibfield  {author} {\bibinfo {author} {\bibfnamefont {A.}~\bibnamefont
  {Singer}}\ and\ \bibinfo {author} {\bibfnamefont {F.~J.}\ \bibnamefont
  {Sigworth}},\ }\bibfield  {title} {\bibinfo {title} {Computational methods
  for single-particle electron cryomicroscopy},\ }\href@noop {} {\bibfield
  {journal} {\bibinfo  {journal} {Annual review of biomedical data science}\
  }\textbf {\bibinfo {volume} {3}},\ \bibinfo {pages} {163} (\bibinfo {year}
  {2020})}\BibitemShut {NoStop}%
\bibitem [{\citenamefont {Tang}\ \emph {et~al.}(2023)\citenamefont {Tang},
  \citenamefont {Zhong}, \citenamefont {Hanson}, \citenamefont {Thiede},\ and\
  \citenamefont {Cossio}}]{tang_conformational_2023}%
  \BibitemOpen
  \bibfield  {author} {\bibinfo {author} {\bibfnamefont {W.~S.}\ \bibnamefont
  {Tang}}, \bibinfo {author} {\bibfnamefont {E.~D.}\ \bibnamefont {Zhong}},
  \bibinfo {author} {\bibfnamefont {S.~M.}\ \bibnamefont {Hanson}}, \bibinfo
  {author} {\bibfnamefont {E.~H.}\ \bibnamefont {Thiede}},\ and\ \bibinfo
  {author} {\bibfnamefont {P.}~\bibnamefont {Cossio}},\ }\bibfield  {title}
  {\bibinfo {title} {Conformational heterogeneity and probability distributions
  from single-particle cryo-electron microscopy},\ }\href@noop {} {\bibfield
  {journal} {\bibinfo  {journal} {Current Opinion in Structural Biology}\
  }\textbf {\bibinfo {volume} {81}},\ \bibinfo {pages} {102626} (\bibinfo
  {year} {2023})}\BibitemShut {NoStop}%
\bibitem [{\citenamefont {Shaw}\ \emph {et~al.}(2014)\citenamefont {Shaw},
  \citenamefont {Grossman}, \citenamefont {Bank}, \citenamefont {Batson},
  \citenamefont {Butts}, \citenamefont {Chao}, \citenamefont {Deneroff},
  \citenamefont {Dror}, \citenamefont {Even}, \citenamefont {Fenton},
  \citenamefont {Forte}, \citenamefont {Gagliardo}, \citenamefont {Gill},
  \citenamefont {Greskamp}, \citenamefont {Ho}, \citenamefont {Ierardi},
  \citenamefont {Iserovich}, \citenamefont {Kuskin}, \citenamefont {Larson},
  \citenamefont {Layman}, \citenamefont {Lee}, \citenamefont {Lerer},
  \citenamefont {Li}, \citenamefont {Killebrew}, \citenamefont {Mackenzie},
  \citenamefont {Mok}, \citenamefont {Moraes}, \citenamefont {Mueller},
  \citenamefont {Nociolo}, \citenamefont {Peticolas}, \citenamefont {Quan},
  \citenamefont {Ramot}, \citenamefont {Salmon}, \citenamefont {Scarpazza},
  \citenamefont {Schafer}, \citenamefont {Siddique}, \citenamefont {Snyder},
  \citenamefont {Spengler}, \citenamefont {Tang}, \citenamefont {Theobald},
  \citenamefont {Toma}, \citenamefont {Towles}, \citenamefont {Vitale},
  \citenamefont {Wang},\ and\ \citenamefont {Young}}]{shaw_anton_2014}%
  \BibitemOpen
  \bibfield  {author} {\bibinfo {author} {\bibfnamefont {D.~E.}\ \bibnamefont
  {Shaw}}, \bibinfo {author} {\bibfnamefont {J.}~\bibnamefont {Grossman}},
  \bibinfo {author} {\bibfnamefont {J.~A.}\ \bibnamefont {Bank}}, \bibinfo
  {author} {\bibfnamefont {B.}~\bibnamefont {Batson}}, \bibinfo {author}
  {\bibfnamefont {J.~A.}\ \bibnamefont {Butts}}, \bibinfo {author}
  {\bibfnamefont {J.~C.}\ \bibnamefont {Chao}}, \bibinfo {author}
  {\bibfnamefont {M.~M.}\ \bibnamefont {Deneroff}}, \bibinfo {author}
  {\bibfnamefont {R.~O.}\ \bibnamefont {Dror}}, \bibinfo {author}
  {\bibfnamefont {A.}~\bibnamefont {Even}}, \bibinfo {author} {\bibfnamefont
  {C.~H.}\ \bibnamefont {Fenton}}, \bibinfo {author} {\bibfnamefont
  {A.}~\bibnamefont {Forte}}, \bibinfo {author} {\bibfnamefont
  {J.}~\bibnamefont {Gagliardo}}, \bibinfo {author} {\bibfnamefont
  {G.}~\bibnamefont {Gill}}, \bibinfo {author} {\bibfnamefont {B.}~\bibnamefont
  {Greskamp}}, \bibinfo {author} {\bibfnamefont {C.~R.}\ \bibnamefont {Ho}},
  \bibinfo {author} {\bibfnamefont {D.~J.}\ \bibnamefont {Ierardi}}, \bibinfo
  {author} {\bibfnamefont {L.}~\bibnamefont {Iserovich}}, \bibinfo {author}
  {\bibfnamefont {J.~S.}\ \bibnamefont {Kuskin}}, \bibinfo {author}
  {\bibfnamefont {R.~H.}\ \bibnamefont {Larson}}, \bibinfo {author}
  {\bibfnamefont {T.}~\bibnamefont {Layman}}, \bibinfo {author} {\bibfnamefont
  {L.-S.}\ \bibnamefont {Lee}}, \bibinfo {author} {\bibfnamefont {A.~K.}\
  \bibnamefont {Lerer}}, \bibinfo {author} {\bibfnamefont {C.}~\bibnamefont
  {Li}}, \bibinfo {author} {\bibfnamefont {D.}~\bibnamefont {Killebrew}},
  \bibinfo {author} {\bibfnamefont {K.~M.}\ \bibnamefont {Mackenzie}}, \bibinfo
  {author} {\bibfnamefont {S.~Y.-H.}\ \bibnamefont {Mok}}, \bibinfo {author}
  {\bibfnamefont {M.~A.}\ \bibnamefont {Moraes}}, \bibinfo {author}
  {\bibfnamefont {R.}~\bibnamefont {Mueller}}, \bibinfo {author} {\bibfnamefont
  {L.~J.}\ \bibnamefont {Nociolo}}, \bibinfo {author} {\bibfnamefont {J.~L.}\
  \bibnamefont {Peticolas}}, \bibinfo {author} {\bibfnamefont {T.}~\bibnamefont
  {Quan}}, \bibinfo {author} {\bibfnamefont {D.}~\bibnamefont {Ramot}},
  \bibinfo {author} {\bibfnamefont {J.~K.}\ \bibnamefont {Salmon}}, \bibinfo
  {author} {\bibfnamefont {D.~P.}\ \bibnamefont {Scarpazza}}, \bibinfo {author}
  {\bibfnamefont {U.~B.}\ \bibnamefont {Schafer}}, \bibinfo {author}
  {\bibfnamefont {N.}~\bibnamefont {Siddique}}, \bibinfo {author}
  {\bibfnamefont {C.~W.}\ \bibnamefont {Snyder}}, \bibinfo {author}
  {\bibfnamefont {J.}~\bibnamefont {Spengler}}, \bibinfo {author}
  {\bibfnamefont {P.~T.~P.}\ \bibnamefont {Tang}}, \bibinfo {author}
  {\bibfnamefont {M.}~\bibnamefont {Theobald}}, \bibinfo {author}
  {\bibfnamefont {H.}~\bibnamefont {Toma}}, \bibinfo {author} {\bibfnamefont
  {B.}~\bibnamefont {Towles}}, \bibinfo {author} {\bibfnamefont
  {B.}~\bibnamefont {Vitale}}, \bibinfo {author} {\bibfnamefont {S.~C.}\
  \bibnamefont {Wang}},\ and\ \bibinfo {author} {\bibfnamefont
  {C.}~\bibnamefont {Young}},\ }\bibfield  {title} {\bibinfo {title} {Anton 2:
  {{Raising}} the bar for performance and programmability in a special-purpose
  molecular dynamics supercomputer},\ }in\ \href@noop {} {\emph {\bibinfo
  {booktitle} {{{SC14}}: {{International Conference}} for {{High Performance
  Computing}}, {{Networking}}, {{Storage}} and {{Analysis}}}}}\ (\bibinfo
  {publisher} {IEEE},\ \bibinfo {address} {New Orleans, LA, USA},\ \bibinfo
  {year} {2014})\ pp.\ \bibinfo {pages} {41--53}\BibitemShut {NoStop}%
\bibitem [{\citenamefont {Shaw}\ \emph {et~al.}(2021)\citenamefont {Shaw},
  \citenamefont {Adams}, \citenamefont {Azaria}, \citenamefont {Bank},
  \citenamefont {Batson}, \citenamefont {Bell}, \citenamefont {Bergdorf},
  \citenamefont {Bhatt}, \citenamefont {Butts}, \citenamefont {Correia},
  \citenamefont {Dirks}, \citenamefont {Dror}, \citenamefont {Eastwood},
  \citenamefont {Edwards}, \citenamefont {Even}, \citenamefont {Feldmann},
  \citenamefont {Fenn}, \citenamefont {Fenton}, \citenamefont {Forte},
  \citenamefont {Gagliardo}, \citenamefont {Gill}, \citenamefont {Gorlatova},
  \citenamefont {Greskamp}, \citenamefont {Grossman}, \citenamefont
  {Gullingsrud}, \citenamefont {Harper}, \citenamefont {Hasenplaugh},
  \citenamefont {Heily}, \citenamefont {Heshmat}, \citenamefont {Hunt},
  \citenamefont {Ierardi}, \citenamefont {Iserovich}, \citenamefont {Jackson},
  \citenamefont {Johnson}, \citenamefont {Kirk}, \citenamefont {Klepeis},
  \citenamefont {Kuskin}, \citenamefont {Mackenzie}, \citenamefont {Mader},
  \citenamefont {McGowen}, \citenamefont {McLaughlin}, \citenamefont {Moraes},
  \citenamefont {Nasr}, \citenamefont {Nociolo}, \citenamefont {O'Donnell},
  \citenamefont {Parker}, \citenamefont {Peticolas}, \citenamefont {Pocina},
  \citenamefont {Predescu}, \citenamefont {Quan}, \citenamefont {Salmon},
  \citenamefont {Schwink}, \citenamefont {Shim}, \citenamefont {Siddique},
  \citenamefont {Spengler}, \citenamefont {Szalay}, \citenamefont {Tabladillo},
  \citenamefont {Tartler}, \citenamefont {Taube}, \citenamefont {Theobald},
  \citenamefont {Towles}, \citenamefont {Vick}, \citenamefont {Wang},
  \citenamefont {Wazlowski}, \citenamefont {Weingarten}, \citenamefont
  {Williams},\ and\ \citenamefont {Yuh}}]{shaw_anton_2021}%
  \BibitemOpen
  \bibfield  {author} {\bibinfo {author} {\bibfnamefont {D.~E.}\ \bibnamefont
  {Shaw}}, \bibinfo {author} {\bibfnamefont {P.~J.}\ \bibnamefont {Adams}},
  \bibinfo {author} {\bibfnamefont {A.}~\bibnamefont {Azaria}}, \bibinfo
  {author} {\bibfnamefont {J.~A.}\ \bibnamefont {Bank}}, \bibinfo {author}
  {\bibfnamefont {B.}~\bibnamefont {Batson}}, \bibinfo {author} {\bibfnamefont
  {A.}~\bibnamefont {Bell}}, \bibinfo {author} {\bibfnamefont {M.}~\bibnamefont
  {Bergdorf}}, \bibinfo {author} {\bibfnamefont {J.}~\bibnamefont {Bhatt}},
  \bibinfo {author} {\bibfnamefont {J.~A.}\ \bibnamefont {Butts}}, \bibinfo
  {author} {\bibfnamefont {T.}~\bibnamefont {Correia}}, \bibinfo {author}
  {\bibfnamefont {R.~M.}\ \bibnamefont {Dirks}}, \bibinfo {author}
  {\bibfnamefont {R.~O.}\ \bibnamefont {Dror}}, \bibinfo {author}
  {\bibfnamefont {M.~P.}\ \bibnamefont {Eastwood}}, \bibinfo {author}
  {\bibfnamefont {B.}~\bibnamefont {Edwards}}, \bibinfo {author} {\bibfnamefont
  {A.}~\bibnamefont {Even}}, \bibinfo {author} {\bibfnamefont {P.}~\bibnamefont
  {Feldmann}}, \bibinfo {author} {\bibfnamefont {M.}~\bibnamefont {Fenn}},
  \bibinfo {author} {\bibfnamefont {C.~H.}\ \bibnamefont {Fenton}}, \bibinfo
  {author} {\bibfnamefont {A.}~\bibnamefont {Forte}}, \bibinfo {author}
  {\bibfnamefont {J.}~\bibnamefont {Gagliardo}}, \bibinfo {author}
  {\bibfnamefont {G.}~\bibnamefont {Gill}}, \bibinfo {author} {\bibfnamefont
  {M.}~\bibnamefont {Gorlatova}}, \bibinfo {author} {\bibfnamefont
  {B.}~\bibnamefont {Greskamp}}, \bibinfo {author} {\bibfnamefont
  {J.}~\bibnamefont {Grossman}}, \bibinfo {author} {\bibfnamefont
  {J.}~\bibnamefont {Gullingsrud}}, \bibinfo {author} {\bibfnamefont
  {A.}~\bibnamefont {Harper}}, \bibinfo {author} {\bibfnamefont
  {W.}~\bibnamefont {Hasenplaugh}}, \bibinfo {author} {\bibfnamefont
  {M.}~\bibnamefont {Heily}}, \bibinfo {author} {\bibfnamefont {B.~C.}\
  \bibnamefont {Heshmat}}, \bibinfo {author} {\bibfnamefont {J.}~\bibnamefont
  {Hunt}}, \bibinfo {author} {\bibfnamefont {D.~J.}\ \bibnamefont {Ierardi}},
  \bibinfo {author} {\bibfnamefont {L.}~\bibnamefont {Iserovich}}, \bibinfo
  {author} {\bibfnamefont {B.~L.}\ \bibnamefont {Jackson}}, \bibinfo {author}
  {\bibfnamefont {N.~P.}\ \bibnamefont {Johnson}}, \bibinfo {author}
  {\bibfnamefont {M.~M.}\ \bibnamefont {Kirk}}, \bibinfo {author}
  {\bibfnamefont {J.~L.}\ \bibnamefont {Klepeis}}, \bibinfo {author}
  {\bibfnamefont {J.~S.}\ \bibnamefont {Kuskin}}, \bibinfo {author}
  {\bibfnamefont {K.~M.}\ \bibnamefont {Mackenzie}}, \bibinfo {author}
  {\bibfnamefont {R.~J.}\ \bibnamefont {Mader}}, \bibinfo {author}
  {\bibfnamefont {R.}~\bibnamefont {McGowen}}, \bibinfo {author} {\bibfnamefont
  {A.}~\bibnamefont {McLaughlin}}, \bibinfo {author} {\bibfnamefont {M.~A.}\
  \bibnamefont {Moraes}}, \bibinfo {author} {\bibfnamefont {M.~H.}\
  \bibnamefont {Nasr}}, \bibinfo {author} {\bibfnamefont {L.~J.}\ \bibnamefont
  {Nociolo}}, \bibinfo {author} {\bibfnamefont {L.}~\bibnamefont {O'Donnell}},
  \bibinfo {author} {\bibfnamefont {A.}~\bibnamefont {Parker}}, \bibinfo
  {author} {\bibfnamefont {J.~L.}\ \bibnamefont {Peticolas}}, \bibinfo {author}
  {\bibfnamefont {G.}~\bibnamefont {Pocina}}, \bibinfo {author} {\bibfnamefont
  {C.}~\bibnamefont {Predescu}}, \bibinfo {author} {\bibfnamefont
  {T.}~\bibnamefont {Quan}}, \bibinfo {author} {\bibfnamefont {J.~K.}\
  \bibnamefont {Salmon}}, \bibinfo {author} {\bibfnamefont {C.}~\bibnamefont
  {Schwink}}, \bibinfo {author} {\bibfnamefont {K.~S.}\ \bibnamefont {Shim}},
  \bibinfo {author} {\bibfnamefont {N.}~\bibnamefont {Siddique}}, \bibinfo
  {author} {\bibfnamefont {J.}~\bibnamefont {Spengler}}, \bibinfo {author}
  {\bibfnamefont {T.}~\bibnamefont {Szalay}}, \bibinfo {author} {\bibfnamefont
  {R.}~\bibnamefont {Tabladillo}}, \bibinfo {author} {\bibfnamefont
  {R.}~\bibnamefont {Tartler}}, \bibinfo {author} {\bibfnamefont {A.~G.}\
  \bibnamefont {Taube}}, \bibinfo {author} {\bibfnamefont {M.}~\bibnamefont
  {Theobald}}, \bibinfo {author} {\bibfnamefont {B.}~\bibnamefont {Towles}},
  \bibinfo {author} {\bibfnamefont {W.}~\bibnamefont {Vick}}, \bibinfo {author}
  {\bibfnamefont {S.~C.}\ \bibnamefont {Wang}}, \bibinfo {author}
  {\bibfnamefont {M.}~\bibnamefont {Wazlowski}}, \bibinfo {author}
  {\bibfnamefont {M.~J.}\ \bibnamefont {Weingarten}}, \bibinfo {author}
  {\bibfnamefont {J.~M.}\ \bibnamefont {Williams}},\ and\ \bibinfo {author}
  {\bibfnamefont {K.~A.}\ \bibnamefont {Yuh}},\ }\bibfield  {title} {\bibinfo
  {title} {Anton 3: twenty microseconds of molecular dynamics simulation before
  lunch},\ }in\ \href@noop {} {\emph {\bibinfo {booktitle} {Proceedings of the
  {{International Conference}} for {{High Performance Computing}},
  {{Networking}}, {{Storage}} and {{Analysis}}}}},\ \bibinfo {series and
  number} {{{SC}} '21}\ (\bibinfo  {publisher} {Association for Computing
  Machinery},\ \bibinfo {address} {New York, NY, USA},\ \bibinfo {year}
  {2021})\ pp.\ \bibinfo {pages} {1--11}\BibitemShut {NoStop}%
\bibitem [{\citenamefont {Shirts}\ and\ \citenamefont
  {Pande}(2000)}]{shirts_computing_2000}%
  \BibitemOpen
  \bibfield  {author} {\bibinfo {author} {\bibfnamefont {M.}~\bibnamefont
  {Shirts}}\ and\ \bibinfo {author} {\bibfnamefont {V.~S.}\ \bibnamefont
  {Pande}},\ }\bibfield  {title} {\bibinfo {title} {{{COMPUTING}}: {{Screen
  Savers}} of the world unite!},\ }\href@noop {} {\bibfield  {journal}
  {\bibinfo  {journal} {Science (New York, N.Y.)}\ }\textbf {\bibinfo {volume}
  {290}},\ \bibinfo {pages} {1903} (\bibinfo {year} {2000})}\BibitemShut
  {NoStop}%
\bibitem [{\citenamefont {Zimmerman}\ \emph {et~al.}(2021)\citenamefont
  {Zimmerman}, \citenamefont {Porter}, \citenamefont {Ward}, \citenamefont
  {Singh}, \citenamefont {Vithani}, \citenamefont {Meller}, \citenamefont
  {Mallimadugula}, \citenamefont {Kuhn}, \citenamefont {Borowsky},
  \citenamefont {Wiewiora}, \citenamefont {Hurley}, \citenamefont {Harbison},
  \citenamefont {Fogarty}, \citenamefont {Coffland}, \citenamefont {Fadda},
  \citenamefont {Voelz}, \citenamefont {Chodera},\ and\ \citenamefont
  {Bowman}}]{zimmerman_sarscov2_2021}%
  \BibitemOpen
  \bibfield  {author} {\bibinfo {author} {\bibfnamefont {M.~I.}\ \bibnamefont
  {Zimmerman}}, \bibinfo {author} {\bibfnamefont {J.~R.}\ \bibnamefont
  {Porter}}, \bibinfo {author} {\bibfnamefont {M.~D.}\ \bibnamefont {Ward}},
  \bibinfo {author} {\bibfnamefont {S.}~\bibnamefont {Singh}}, \bibinfo
  {author} {\bibfnamefont {N.}~\bibnamefont {Vithani}}, \bibinfo {author}
  {\bibfnamefont {A.}~\bibnamefont {Meller}}, \bibinfo {author} {\bibfnamefont
  {U.~L.}\ \bibnamefont {Mallimadugula}}, \bibinfo {author} {\bibfnamefont
  {C.~E.}\ \bibnamefont {Kuhn}}, \bibinfo {author} {\bibfnamefont {J.~H.}\
  \bibnamefont {Borowsky}}, \bibinfo {author} {\bibfnamefont {R.~P.}\
  \bibnamefont {Wiewiora}}, \bibinfo {author} {\bibfnamefont {M.~F.~D.}\
  \bibnamefont {Hurley}}, \bibinfo {author} {\bibfnamefont {A.~M.}\
  \bibnamefont {Harbison}}, \bibinfo {author} {\bibfnamefont {C.~A.}\
  \bibnamefont {Fogarty}}, \bibinfo {author} {\bibfnamefont {J.~E.}\
  \bibnamefont {Coffland}}, \bibinfo {author} {\bibfnamefont {E.}~\bibnamefont
  {Fadda}}, \bibinfo {author} {\bibfnamefont {V.~A.}\ \bibnamefont {Voelz}},
  \bibinfo {author} {\bibfnamefont {J.~D.}\ \bibnamefont {Chodera}},\ and\
  \bibinfo {author} {\bibfnamefont {G.~R.}\ \bibnamefont {Bowman}},\ }\bibfield
   {title} {\bibinfo {title} {{{SARS-CoV-2}} simulations go exascale to predict
  dramatic spike opening and cryptic pockets across the proteome},\ }\href@noop
  {} {\bibfield  {journal} {\bibinfo  {journal} {Nature Chemistry}\ }\textbf
  {\bibinfo {volume} {13}},\ \bibinfo {pages} {651} (\bibinfo {year}
  {2021})}\BibitemShut {NoStop}%
\bibitem [{\citenamefont {Ma}\ and\ \citenamefont
  {Dinner}(2005)}]{ma_automatic_2005}%
  \BibitemOpen
  \bibfield  {author} {\bibinfo {author} {\bibfnamefont {A.}~\bibnamefont
  {Ma}}\ and\ \bibinfo {author} {\bibfnamefont {A.~R.}\ \bibnamefont
  {Dinner}},\ }\bibfield  {title} {\bibinfo {title} {Automatic method for
  identifying reaction coordinates in complex systems},\ }\href
  {https://doi.org/10.1021/jp045546c} {\bibfield  {journal} {\bibinfo
  {journal} {Journal of Physical Chemistry B}\ }\textbf {\bibinfo {volume}
  {109}},\ \bibinfo {pages} {6769} (\bibinfo {year} {2005})}\BibitemShut
  {NoStop}%
\bibitem [{\citenamefont {Rohrdanz}\ \emph {et~al.}(2011)\citenamefont
  {Rohrdanz}, \citenamefont {Zheng}, \citenamefont {Maggioni},\ and\
  \citenamefont {Clementi}}]{rohrdanz_determination_2011}%
  \BibitemOpen
  \bibfield  {author} {\bibinfo {author} {\bibfnamefont {M.~A.}\ \bibnamefont
  {Rohrdanz}}, \bibinfo {author} {\bibfnamefont {W.}~\bibnamefont {Zheng}},
  \bibinfo {author} {\bibfnamefont {M.}~\bibnamefont {Maggioni}},\ and\
  \bibinfo {author} {\bibfnamefont {C.}~\bibnamefont {Clementi}},\ }\bibfield
  {title} {\bibinfo {title} {Determination of reaction coordinates via locally
  scaled diffusion map},\ }\href {https://doi.org/10.1063/1.3569857} {\bibfield
   {journal} {\bibinfo  {journal} {Journal of Chemical Physics}\ }\textbf
  {\bibinfo {volume} {134}},\ \bibinfo {pages} {124116} (\bibinfo {year}
  {2011})}\BibitemShut {NoStop}%
\bibitem [{\citenamefont {Mardt}\ \emph {et~al.}(2018)\citenamefont {Mardt},
  \citenamefont {Pasquali}, \citenamefont {Wu},\ and\ \citenamefont
  {Noé}}]{mardt_vampnets_2018}%
  \BibitemOpen
  \bibfield  {author} {\bibinfo {author} {\bibfnamefont {A.}~\bibnamefont
  {Mardt}}, \bibinfo {author} {\bibfnamefont {L.}~\bibnamefont {Pasquali}},
  \bibinfo {author} {\bibfnamefont {H.}~\bibnamefont {Wu}},\ and\ \bibinfo
  {author} {\bibfnamefont {F.}~\bibnamefont {Noé}},\ }\bibfield  {title}
  {\bibinfo {title} {{VAMPnets} for deep learning of molecular kinetics},\
  }\href {https://doi.org/10.1038/s41467-017-02388-1} {\bibfield  {journal}
  {\bibinfo  {journal} {Nature Communications}\ }\textbf {\bibinfo {volume}
  {9}},\ \bibinfo {pages} {5} (\bibinfo {year} {2018})}\BibitemShut {NoStop}%
\bibitem [{\citenamefont {Wang}\ and\ \citenamefont
  {Tiwary}(2021)}]{wang2021state}%
  \BibitemOpen
  \bibfield  {author} {\bibinfo {author} {\bibfnamefont {D.}~\bibnamefont
  {Wang}}\ and\ \bibinfo {author} {\bibfnamefont {P.}~\bibnamefont {Tiwary}},\
  }\bibfield  {title} {\bibinfo {title} {State predictive information
  bottleneck},\ }\href@noop {} {\bibfield  {journal} {\bibinfo  {journal}
  {Journal of Chemical Physics}\ }\textbf {\bibinfo {volume} {154}},\ \bibinfo
  {pages} {134111} (\bibinfo {year} {2021})}\BibitemShut {NoStop}%
\bibitem [{\citenamefont {Chen}\ \emph {et~al.}(2023)\citenamefont {Chen},
  \citenamefont {Roux},\ and\ \citenamefont {Chipot}}]{chen_discovering_2023}%
  \BibitemOpen
  \bibfield  {author} {\bibinfo {author} {\bibfnamefont {H.}~\bibnamefont
  {Chen}}, \bibinfo {author} {\bibfnamefont {B.}~\bibnamefont {Roux}},\ and\
  \bibinfo {author} {\bibfnamefont {C.}~\bibnamefont {Chipot}},\ }\bibfield
  {title} {\bibinfo {title} {Discovering reaction pathways, slow variables, and
  committor probabilities with machine learning},\ }\href
  {https://doi.org/10.1021/acs.jctc.3c00028} {\bibfield  {journal} {\bibinfo
  {journal} {Journal of Chemical Theory and Computation}\ }\textbf {\bibinfo
  {volume} {19}},\ \bibinfo {pages} {4414} (\bibinfo {year}
  {2023})}\BibitemShut {NoStop}%
\bibitem [{\citenamefont {Bonati}\ \emph {et~al.}(2023)\citenamefont {Bonati},
  \citenamefont {Trizio}, \citenamefont {Rizzi},\ and\ \citenamefont
  {Parrinello}}]{bonati2023unified}%
  \BibitemOpen
  \bibfield  {author} {\bibinfo {author} {\bibfnamefont {L.}~\bibnamefont
  {Bonati}}, \bibinfo {author} {\bibfnamefont {E.}~\bibnamefont {Trizio}},
  \bibinfo {author} {\bibfnamefont {A.}~\bibnamefont {Rizzi}},\ and\ \bibinfo
  {author} {\bibfnamefont {M.}~\bibnamefont {Parrinello}},\ }\bibfield  {title}
  {\bibinfo {title} {A unified framework for machine learning collective
  variables for enhanced sampling simulations: mlcolvar},\ }\href@noop {}
  {\bibfield  {journal} {\bibinfo  {journal} {The Journal of Chemical Physics}\
  }\textbf {\bibinfo {volume} {159}} (\bibinfo {year} {2023})}\BibitemShut
  {NoStop}%
\bibitem [{\citenamefont {Zhang}\ \emph {et~al.}(2024)\citenamefont {Zhang},
  \citenamefont {Bonati}, \citenamefont {Trizio}, \citenamefont {Zhang},
  \citenamefont {Kang}, \citenamefont {Hou},\ and\ \citenamefont
  {Parrinello}}]{zhang2024descriptor}%
  \BibitemOpen
  \bibfield  {author} {\bibinfo {author} {\bibfnamefont {J.}~\bibnamefont
  {Zhang}}, \bibinfo {author} {\bibfnamefont {L.}~\bibnamefont {Bonati}},
  \bibinfo {author} {\bibfnamefont {E.}~\bibnamefont {Trizio}}, \bibinfo
  {author} {\bibfnamefont {O.}~\bibnamefont {Zhang}}, \bibinfo {author}
  {\bibfnamefont {Y.}~\bibnamefont {Kang}}, \bibinfo {author} {\bibfnamefont
  {T.}~\bibnamefont {Hou}},\ and\ \bibinfo {author} {\bibfnamefont
  {M.}~\bibnamefont {Parrinello}},\ }\bibfield  {title} {\bibinfo {title}
  {Descriptor-free collective variables from geometric graph neural networks},\
  }\href@noop {} {\bibfield  {journal} {\bibinfo  {journal} {Journal of
  Chemical Theory and Computation}\ }\textbf {\bibinfo {volume} {20}},\
  \bibinfo {pages} {10787} (\bibinfo {year} {2024})}\BibitemShut {NoStop}%
\bibitem [{\citenamefont {Zou}\ \emph {et~al.}(2025)\citenamefont {Zou},
  \citenamefont {Wang},\ and\ \citenamefont {Tiwary}}]{zou2025graph}%
  \BibitemOpen
  \bibfield  {author} {\bibinfo {author} {\bibfnamefont {Z.}~\bibnamefont
  {Zou}}, \bibinfo {author} {\bibfnamefont {D.}~\bibnamefont {Wang}},\ and\
  \bibinfo {author} {\bibfnamefont {P.}~\bibnamefont {Tiwary}},\ }\bibfield
  {title} {\bibinfo {title} {A graph neural network-state predictive
  information bottleneck (gnn-spib) approach for learning molecular
  thermodynamics and kinetics},\ }\href@noop {} {\bibfield  {journal} {\bibinfo
   {journal} {Digital Discovery}\ }\textbf {\bibinfo {volume} {4}},\ \bibinfo
  {pages} {211} (\bibinfo {year} {2025})}\BibitemShut {NoStop}%
\bibitem [{\citenamefont {Gilmer}\ \emph {et~al.}(2017)\citenamefont {Gilmer},
  \citenamefont {Schoenholz}, \citenamefont {Riley}, \citenamefont {Vinyals},\
  and\ \citenamefont {Dahl}}]{gilmer2017neural}%
  \BibitemOpen
  \bibfield  {author} {\bibinfo {author} {\bibfnamefont {J.}~\bibnamefont
  {Gilmer}}, \bibinfo {author} {\bibfnamefont {S.~S.}\ \bibnamefont
  {Schoenholz}}, \bibinfo {author} {\bibfnamefont {P.~F.}\ \bibnamefont
  {Riley}}, \bibinfo {author} {\bibfnamefont {O.}~\bibnamefont {Vinyals}},\
  and\ \bibinfo {author} {\bibfnamefont {G.~E.}\ \bibnamefont {Dahl}},\
  }\bibfield  {title} {\bibinfo {title} {Neural message passing for quantum
  chemistry},\ }in\ \href@noop {} {\emph {\bibinfo {booktitle} {International
  Conference on Machine Learning}}}\ (\bibinfo {organization} {PMLR},\ \bibinfo
  {year} {2017})\ pp.\ \bibinfo {pages} {1263--1272}\BibitemShut {NoStop}%
\bibitem [{\citenamefont {Jamasb}\ \emph {et~al.}(2023)\citenamefont {Jamasb},
  \citenamefont {Morehead}, \citenamefont {Zhang}, \citenamefont {Joshi},
  \citenamefont {Didi}, \citenamefont {Mathis}, \citenamefont {Harris},
  \citenamefont {Tang}, \citenamefont {Cheng}, \citenamefont {Lio},\ and\
  \citenamefont {Blundell}}]{jamasb_evaluating_2023}%
  \BibitemOpen
  \bibfield  {author} {\bibinfo {author} {\bibfnamefont {A.~R.}\ \bibnamefont
  {Jamasb}}, \bibinfo {author} {\bibfnamefont {A.}~\bibnamefont {Morehead}},
  \bibinfo {author} {\bibfnamefont {Z.}~\bibnamefont {Zhang}}, \bibinfo
  {author} {\bibfnamefont {C.~K.}\ \bibnamefont {Joshi}}, \bibinfo {author}
  {\bibfnamefont {K.}~\bibnamefont {Didi}}, \bibinfo {author} {\bibfnamefont
  {S.~V.}\ \bibnamefont {Mathis}}, \bibinfo {author} {\bibfnamefont
  {C.}~\bibnamefont {Harris}}, \bibinfo {author} {\bibfnamefont
  {J.}~\bibnamefont {Tang}}, \bibinfo {author} {\bibfnamefont {J.}~\bibnamefont
  {Cheng}}, \bibinfo {author} {\bibfnamefont {P.}~\bibnamefont {Lio}},\ and\
  \bibinfo {author} {\bibfnamefont {T.~L.}\ \bibnamefont {Blundell}},\
  }\bibfield  {title} {\bibinfo {title} {Evaluating representation learning on
  the protein structure universe},\ }in\ \href@noop {} {\emph {\bibinfo
  {booktitle} {The {{Twelfth International Conference}} on {{Learning
  Representations}}}}}\ (\bibinfo {year} {2023})\BibitemShut {NoStop}%
\bibitem [{\citenamefont {Xie}\ \emph {et~al.}(2019)\citenamefont {Xie},
  \citenamefont {France-Lanord}, \citenamefont {Wang}, \citenamefont
  {Shao-Horn},\ and\ \citenamefont {Grossman}}]{xie2019graph}%
  \BibitemOpen
  \bibfield  {author} {\bibinfo {author} {\bibfnamefont {T.}~\bibnamefont
  {Xie}}, \bibinfo {author} {\bibfnamefont {A.}~\bibnamefont {France-Lanord}},
  \bibinfo {author} {\bibfnamefont {Y.}~\bibnamefont {Wang}}, \bibinfo {author}
  {\bibfnamefont {Y.}~\bibnamefont {Shao-Horn}},\ and\ \bibinfo {author}
  {\bibfnamefont {J.~C.}\ \bibnamefont {Grossman}},\ }\bibfield  {title}
  {\bibinfo {title} {Graph dynamical networks for unsupervised learning of
  atomic scale dynamics in materials},\ }\href@noop {} {\bibfield  {journal}
  {\bibinfo  {journal} {Nature Communications}\ }\textbf {\bibinfo {volume}
  {10}},\ \bibinfo {pages} {2667} (\bibinfo {year} {2019})}\BibitemShut
  {NoStop}%
\bibitem [{\citenamefont {Patel}\ \emph {et~al.}(2024)\citenamefont {Patel},
  \citenamefont {Sinha},\ and\ \citenamefont {Palermo}}]{patel2024graph}%
  \BibitemOpen
  \bibfield  {author} {\bibinfo {author} {\bibfnamefont {A.~C.}\ \bibnamefont
  {Patel}}, \bibinfo {author} {\bibfnamefont {S.}~\bibnamefont {Sinha}},\ and\
  \bibinfo {author} {\bibfnamefont {G.}~\bibnamefont {Palermo}},\ }\bibfield
  {title} {\bibinfo {title} {Graph theory approaches for molecular dynamics
  simulations},\ }\href@noop {} {\bibfield  {journal} {\bibinfo  {journal}
  {Quarterly reviews of biophysics}\ }\textbf {\bibinfo {volume} {57}},\
  \bibinfo {pages} {e15} (\bibinfo {year} {2024})}\BibitemShut {NoStop}%
\bibitem [{\citenamefont {Arredondo}\ \emph {et~al.}(2025)\citenamefont
  {Arredondo}, \citenamefont {Tang}, \citenamefont {Talmazan}, \citenamefont
  {Meg{\'\i}as}, \citenamefont {Chen},\ and\ \citenamefont
  {Chipot}}]{arredondo2025atoms}%
  \BibitemOpen
  \bibfield  {author} {\bibinfo {author} {\bibfnamefont {S.~C.}\ \bibnamefont
  {Arredondo}}, \bibinfo {author} {\bibfnamefont {C.}~\bibnamefont {Tang}},
  \bibinfo {author} {\bibfnamefont {R.~A.}\ \bibnamefont {Talmazan}}, \bibinfo
  {author} {\bibfnamefont {A.}~\bibnamefont {Meg{\'\i}as}}, \bibinfo {author}
  {\bibfnamefont {C.~G.}\ \bibnamefont {Chen}},\ and\ \bibinfo {author}
  {\bibfnamefont {C.}~\bibnamefont {Chipot}},\ }\bibfield  {title} {\bibinfo
  {title} {From atoms to dynamics: Learning the committor without collective
  variables},\ }\href@noop {} {\bibfield  {journal} {\bibinfo  {journal} {arXiv
  preprint arXiv:2507.17700}\ } (\bibinfo {year} {2025})}\BibitemShut {NoStop}%
\bibitem [{\citenamefont {Anderson}\ \emph {et~al.}(2019)\citenamefont
  {Anderson}, \citenamefont {Hy},\ and\ \citenamefont
  {Kondor}}]{anderson2019cormorant}%
  \BibitemOpen
  \bibfield  {author} {\bibinfo {author} {\bibfnamefont {B.}~\bibnamefont
  {Anderson}}, \bibinfo {author} {\bibfnamefont {T.-S.}\ \bibnamefont {Hy}},\
  and\ \bibinfo {author} {\bibfnamefont {R.}~\bibnamefont {Kondor}},\
  }\bibfield  {title} {\bibinfo {title} {Cormorant: Covariant molecular neural
  networks},\ }in\ \href@noop {} {\emph {\bibinfo {booktitle} {Proceedings of
  the 33rd {{International Conference}} on {{Neural Information Processing
  Systems}}}}}\ (\bibinfo  {publisher} {Curran Associates Inc.},\ \bibinfo
  {address} {Red Hook, NY, USA},\ \bibinfo {year} {2019})\ pp.\ \bibinfo
  {pages} {14537--14546}\BibitemShut {NoStop}%
\bibitem [{\citenamefont {Wang}\ \emph
  {et~al.}(2024{\natexlab{a}})\citenamefont {Wang}, \citenamefont {Wang},
  \citenamefont {Li}, \citenamefont {He}, \citenamefont {Li}, \citenamefont
  {Wang}, \citenamefont {Zheng}, \citenamefont {Shao},\ and\ \citenamefont
  {Liu}}]{wang2024enhancing}%
  \BibitemOpen
  \bibfield  {author} {\bibinfo {author} {\bibfnamefont {Y.}~\bibnamefont
  {Wang}}, \bibinfo {author} {\bibfnamefont {T.}~\bibnamefont {Wang}}, \bibinfo
  {author} {\bibfnamefont {S.}~\bibnamefont {Li}}, \bibinfo {author}
  {\bibfnamefont {X.}~\bibnamefont {He}}, \bibinfo {author} {\bibfnamefont
  {M.}~\bibnamefont {Li}}, \bibinfo {author} {\bibfnamefont {Z.}~\bibnamefont
  {Wang}}, \bibinfo {author} {\bibfnamefont {N.}~\bibnamefont {Zheng}},
  \bibinfo {author} {\bibfnamefont {B.}~\bibnamefont {Shao}},\ and\ \bibinfo
  {author} {\bibfnamefont {T.-Y.}\ \bibnamefont {Liu}},\ }\bibfield  {title}
  {\bibinfo {title} {Enhancing geometric representations for molecules with
  equivariant vector-scalar interactive message passing},\ }\href@noop {}
  {\bibfield  {journal} {\bibinfo  {journal} {Nature Communications}\ }\textbf
  {\bibinfo {volume} {15}},\ \bibinfo {pages} {313} (\bibinfo {year}
  {2024}{\natexlab{a}})}\BibitemShut {NoStop}%
\bibitem [{\citenamefont {Pengmei}\ \emph {et~al.}(2025)\citenamefont
  {Pengmei}, \citenamefont {Lorpaiboon}, \citenamefont {Guo}, \citenamefont
  {Weare},\ and\ \citenamefont {Dinner}}]{pengmei2025using}%
  \BibitemOpen
  \bibfield  {author} {\bibinfo {author} {\bibfnamefont {Z.}~\bibnamefont
  {Pengmei}}, \bibinfo {author} {\bibfnamefont {C.}~\bibnamefont {Lorpaiboon}},
  \bibinfo {author} {\bibfnamefont {S.~C.}\ \bibnamefont {Guo}}, \bibinfo
  {author} {\bibfnamefont {J.}~\bibnamefont {Weare}},\ and\ \bibinfo {author}
  {\bibfnamefont {A.~R.}\ \bibnamefont {Dinner}},\ }\bibfield  {title}
  {\bibinfo {title} {Using pretrained graph neural networks with token mixers
  as geometric featurizers for conformational dynamics},\ }\href@noop {}
  {\bibfield  {journal} {\bibinfo  {journal} {The Journal of Chemical Physics}\
  }\textbf {\bibinfo {volume} {162}} (\bibinfo {year} {2025})}\BibitemShut
  {NoStop}%
\bibitem [{\citenamefont {Ghorbani}\ \emph {et~al.}(2022)\citenamefont
  {Ghorbani}, \citenamefont {Prasad}, \citenamefont {Klauda},\ and\
  \citenamefont {Brooks}}]{ghorbani2022graphvampnet}%
  \BibitemOpen
  \bibfield  {author} {\bibinfo {author} {\bibfnamefont {M.}~\bibnamefont
  {Ghorbani}}, \bibinfo {author} {\bibfnamefont {S.}~\bibnamefont {Prasad}},
  \bibinfo {author} {\bibfnamefont {J.~B.}\ \bibnamefont {Klauda}},\ and\
  \bibinfo {author} {\bibfnamefont {B.~R.}\ \bibnamefont {Brooks}},\ }\bibfield
   {title} {\bibinfo {title} {{GraphVAMPNet}, using graph neural networks and
  variational approach to {Markov} processes for dynamical modeling of
  biomolecules},\ }\href@noop {} {\bibfield  {journal} {\bibinfo  {journal}
  {Journal of Chemical Physics}\ }\textbf {\bibinfo {volume} {156}},\ \bibinfo
  {pages} {184103} (\bibinfo {year} {2022})}\BibitemShut {NoStop}%
\bibitem [{\citenamefont {Liu}\ \emph {et~al.}(2023)\citenamefont {Liu},
  \citenamefont {Xue}, \citenamefont {Qiu}, \citenamefont {Konovalov},
  \citenamefont {O’Connor},\ and\ \citenamefont
  {Huang}}]{liu2023graphvampnets}%
  \BibitemOpen
  \bibfield  {author} {\bibinfo {author} {\bibfnamefont {B.}~\bibnamefont
  {Liu}}, \bibinfo {author} {\bibfnamefont {M.}~\bibnamefont {Xue}}, \bibinfo
  {author} {\bibfnamefont {Y.}~\bibnamefont {Qiu}}, \bibinfo {author}
  {\bibfnamefont {K.~A.}\ \bibnamefont {Konovalov}}, \bibinfo {author}
  {\bibfnamefont {M.~S.}\ \bibnamefont {O’Connor}},\ and\ \bibinfo {author}
  {\bibfnamefont {X.}~\bibnamefont {Huang}},\ }\bibfield  {title} {\bibinfo
  {title} {{GraphVAMPnets} for uncovering slow collective variables of
  self-assembly dynamics},\ }\href@noop {} {\bibfield  {journal} {\bibinfo
  {journal} {Journal of Chemical Physics}\ }\textbf {\bibinfo {volume} {159}},\
  \bibinfo {pages} {094901} (\bibinfo {year} {2023})}\BibitemShut {NoStop}%
\bibitem [{\citenamefont {Bolya}\ \emph {et~al.}(2022)\citenamefont {Bolya},
  \citenamefont {Fu}, \citenamefont {Dai}, \citenamefont {Zhang}, \citenamefont
  {Feichtenhofer},\ and\ \citenamefont {Hoffman}}]{bolya2022token}%
  \BibitemOpen
  \bibfield  {author} {\bibinfo {author} {\bibfnamefont {D.}~\bibnamefont
  {Bolya}}, \bibinfo {author} {\bibfnamefont {C.-Y.}\ \bibnamefont {Fu}},
  \bibinfo {author} {\bibfnamefont {X.}~\bibnamefont {Dai}}, \bibinfo {author}
  {\bibfnamefont {P.}~\bibnamefont {Zhang}}, \bibinfo {author} {\bibfnamefont
  {C.}~\bibnamefont {Feichtenhofer}},\ and\ \bibinfo {author} {\bibfnamefont
  {J.}~\bibnamefont {Hoffman}},\ }\bibfield  {title} {\bibinfo {title} {Token
  merging: Your {ViT} but faster},\ }\href@noop {} {\bibfield  {journal}
  {\bibinfo  {journal} {arXiv preprint arXiv:2210.09461}\ } (\bibinfo {year}
  {2022})}\BibitemShut {NoStop}%
\bibitem [{\citenamefont {Dao}\ \emph {et~al.}(2022)\citenamefont {Dao},
  \citenamefont {Fu}, \citenamefont {Ermon}, \citenamefont {Rudra},\ and\
  \citenamefont {R{\'e}}}]{dao2022flashattention}%
  \BibitemOpen
  \bibfield  {author} {\bibinfo {author} {\bibfnamefont {T.}~\bibnamefont
  {Dao}}, \bibinfo {author} {\bibfnamefont {D.}~\bibnamefont {Fu}}, \bibinfo
  {author} {\bibfnamefont {S.}~\bibnamefont {Ermon}}, \bibinfo {author}
  {\bibfnamefont {A.}~\bibnamefont {Rudra}},\ and\ \bibinfo {author}
  {\bibfnamefont {C.}~\bibnamefont {R{\'e}}},\ }\bibfield  {title} {\bibinfo
  {title} {{FlashAttention}: Fast and memory-efficient exact attention with
  {IO}-awareness},\ }\href@noop {} {\bibfield  {journal} {\bibinfo  {journal}
  {Advances in Neural Information Processing Systems}\ }\textbf {\bibinfo
  {volume} {35}},\ \bibinfo {pages} {16344} (\bibinfo {year}
  {2022})}\BibitemShut {NoStop}%
\bibitem [{\citenamefont {Fuson}\ \emph {et~al.}(2007)\citenamefont {Fuson},
  \citenamefont {Montes}, \citenamefont {Robert},\ and\ \citenamefont
  {Sutton}}]{fuson_structure_2007}%
  \BibitemOpen
  \bibfield  {author} {\bibinfo {author} {\bibfnamefont {K.~L.}\ \bibnamefont
  {Fuson}}, \bibinfo {author} {\bibfnamefont {M.}~\bibnamefont {Montes}},
  \bibinfo {author} {\bibfnamefont {J.~J.}\ \bibnamefont {Robert}},\ and\
  \bibinfo {author} {\bibfnamefont {R.~B.}\ \bibnamefont {Sutton}},\ }\bibfield
   {title} {\bibinfo {title} {Structure of human synaptotagmin 1 {{C2AB}} in
  the absence of {Ca\textsuperscript{2+}} reveals a novel domain association},\
  }\href@noop {} {\bibfield  {journal} {\bibinfo  {journal} {Biochemistry}\
  }\textbf {\bibinfo {volume} {46}},\ \bibinfo {pages} {13041} (\bibinfo {year}
  {2007})}\BibitemShut {NoStop}%
\bibitem [{\citenamefont {M{\"u}ller}\ \emph {et~al.}(1996)\citenamefont
  {M{\"u}ller}, \citenamefont {Schlauderer}, \citenamefont {Reinstein},\ and\
  \citenamefont {Schulz}}]{muller_adenylate_1996}%
  \BibitemOpen
  \bibfield  {author} {\bibinfo {author} {\bibfnamefont {{\relax
  CW}.}~\bibnamefont {M{\"u}ller}}, \bibinfo {author} {\bibfnamefont {{\relax
  GJ}.}~\bibnamefont {Schlauderer}}, \bibinfo {author} {\bibfnamefont
  {J.}~\bibnamefont {Reinstein}},\ and\ \bibinfo {author} {\bibfnamefont
  {{\relax GE}.}~\bibnamefont {Schulz}},\ }\bibfield  {title} {\bibinfo {title}
  {Adenylate kinase motions during catalysis: an energetic counterweight
  balancing substrate binding},\ }\href@noop {} {\bibfield  {journal} {\bibinfo
   {journal} {Structure}\ }\textbf {\bibinfo {volume} {4}},\ \bibinfo {pages}
  {147} (\bibinfo {year} {1996})}\BibitemShut {NoStop}%
\bibitem [{\citenamefont {Chen}\ \emph {et~al.}(2022)\citenamefont {Chen},
  \citenamefont {Wang}, \citenamefont {Malone}, \citenamefont {Llewellyn},
  \citenamefont {Pechersky}, \citenamefont {Maruthi}, \citenamefont {Eng},
  \citenamefont {Perry}, \citenamefont {Campbell}, \citenamefont {Shaw},\ and\
  \citenamefont {Darst}}]{chen_ensemble_2022}%
  \BibitemOpen
  \bibfield  {author} {\bibinfo {author} {\bibfnamefont {J.}~\bibnamefont
  {Chen}}, \bibinfo {author} {\bibfnamefont {Q.}~\bibnamefont {Wang}}, \bibinfo
  {author} {\bibfnamefont {B.}~\bibnamefont {Malone}}, \bibinfo {author}
  {\bibfnamefont {E.}~\bibnamefont {Llewellyn}}, \bibinfo {author}
  {\bibfnamefont {Y.}~\bibnamefont {Pechersky}}, \bibinfo {author}
  {\bibfnamefont {K.}~\bibnamefont {Maruthi}}, \bibinfo {author} {\bibfnamefont
  {E.~T.}\ \bibnamefont {Eng}}, \bibinfo {author} {\bibfnamefont {J.~K.}\
  \bibnamefont {Perry}}, \bibinfo {author} {\bibfnamefont {E.~A.}\ \bibnamefont
  {Campbell}}, \bibinfo {author} {\bibfnamefont {D.~E.}\ \bibnamefont {Shaw}},\
  and\ \bibinfo {author} {\bibfnamefont {S.~A.}\ \bibnamefont {Darst}},\
  }\bibfield  {title} {\bibinfo {title} {Ensemble cryo-{{EM}} reveals
  conformational states of the nsp13 helicase in the {{SARS-CoV-2}} helicase
  replication--transcription complex},\ }\href@noop {} {\bibfield  {journal}
  {\bibinfo  {journal} {Nature Structural \& Molecular Biology}\ }\textbf
  {\bibinfo {volume} {29}},\ \bibinfo {pages} {250} (\bibinfo {year}
  {2022})}\BibitemShut {NoStop}%
\bibitem [{\citenamefont {Zaidi}\ \emph {et~al.}(2022)\citenamefont {Zaidi},
  \citenamefont {Schaarschmidt}, \citenamefont {Martens}, \citenamefont {Kim},
  \citenamefont {Teh}, \citenamefont {{Sanchez-Gonzalez}}, \citenamefont
  {Battaglia}, \citenamefont {Pascanu},\ and\ \citenamefont
  {Godwin}}]{zaidi2022pre}%
  \BibitemOpen
  \bibfield  {author} {\bibinfo {author} {\bibfnamefont {S.}~\bibnamefont
  {Zaidi}}, \bibinfo {author} {\bibfnamefont {M.}~\bibnamefont
  {Schaarschmidt}}, \bibinfo {author} {\bibfnamefont {J.}~\bibnamefont
  {Martens}}, \bibinfo {author} {\bibfnamefont {H.}~\bibnamefont {Kim}},
  \bibinfo {author} {\bibfnamefont {Y.~W.}\ \bibnamefont {Teh}}, \bibinfo
  {author} {\bibfnamefont {A.}~\bibnamefont {{Sanchez-Gonzalez}}}, \bibinfo
  {author} {\bibfnamefont {P.}~\bibnamefont {Battaglia}}, \bibinfo {author}
  {\bibfnamefont {R.}~\bibnamefont {Pascanu}},\ and\ \bibinfo {author}
  {\bibfnamefont {J.}~\bibnamefont {Godwin}},\ }\bibfield  {title} {\bibinfo
  {title} {Pre-training via denoising for molecular property prediction},\ }in\
  \href@noop {} {\emph {\bibinfo {booktitle} {The {{Eleventh International
  Conference}} on {{Learning Representations}}}}}\ (\bibinfo {year}
  {2022})\BibitemShut {NoStop}%
\bibitem [{\citenamefont {Pengmei}\ \emph
  {et~al.}(2024{\natexlab{a}})\citenamefont {Pengmei}, \citenamefont
  {Lorpaiboon}, \citenamefont {Guo}, \citenamefont {Weare},\ and\ \citenamefont
  {Dinner}}]{pengmei2024geom2vec}%
  \BibitemOpen
  \bibfield  {author} {\bibinfo {author} {\bibfnamefont {Z.}~\bibnamefont
  {Pengmei}}, \bibinfo {author} {\bibfnamefont {C.}~\bibnamefont {Lorpaiboon}},
  \bibinfo {author} {\bibfnamefont {S.~C.}\ \bibnamefont {Guo}}, \bibinfo
  {author} {\bibfnamefont {J.}~\bibnamefont {Weare}},\ and\ \bibinfo {author}
  {\bibfnamefont {A.~R.}\ \bibnamefont {Dinner}},\ }\bibfield  {title}
  {\bibinfo {title} {geom2vec: pretrained {GNNs} as geometric featurizers for
  conformational dynamics},\ }\href@noop {} {\bibfield  {journal} {\bibinfo
  {journal} {arXiv preprint arXiv:2409.19838}\ } (\bibinfo {year}
  {2024}{\natexlab{a}})}\BibitemShut {NoStop}%
\bibitem [{\citenamefont {Pengmei}\ \emph
  {et~al.}(2024{\natexlab{b}})\citenamefont {Pengmei}, \citenamefont {Shen},
  \citenamefont {Wang}, \citenamefont {Collins},\ and\ \citenamefont
  {Rangwala}}]{pengmei2024pushinglimitsallatomgeometric}%
  \BibitemOpen
  \bibfield  {author} {\bibinfo {author} {\bibfnamefont {Z.}~\bibnamefont
  {Pengmei}}, \bibinfo {author} {\bibfnamefont {Z.}~\bibnamefont {Shen}},
  \bibinfo {author} {\bibfnamefont {Z.}~\bibnamefont {Wang}}, \bibinfo {author}
  {\bibfnamefont {M.}~\bibnamefont {Collins}},\ and\ \bibinfo {author}
  {\bibfnamefont {H.}~\bibnamefont {Rangwala}},\ }\bibfield  {title} {\bibinfo
  {title} {Pushing the limits of all-atom geometric graph neural networks:
  Pre-training, scaling and zero-shot transfer},\ }\href@noop {} {\bibfield
  {journal} {\bibinfo  {journal} {arXiv preprint arXiv:2410.21683}\ } (\bibinfo
  {year} {2024}{\natexlab{b}})}\BibitemShut {NoStop}%
\bibitem [{\citenamefont {Vaswani}\ \emph {et~al.}(2017)\citenamefont
  {Vaswani}, \citenamefont {Shazeer}, \citenamefont {Parmar}, \citenamefont
  {Uszkoreit}, \citenamefont {Jones}, \citenamefont {Gomez}, \citenamefont
  {Kaiser},\ and\ \citenamefont {Polosukhin}}]{vaswani2017attention}%
  \BibitemOpen
  \bibfield  {author} {\bibinfo {author} {\bibfnamefont {A.}~\bibnamefont
  {Vaswani}}, \bibinfo {author} {\bibfnamefont {N.}~\bibnamefont {Shazeer}},
  \bibinfo {author} {\bibfnamefont {N.}~\bibnamefont {Parmar}}, \bibinfo
  {author} {\bibfnamefont {J.}~\bibnamefont {Uszkoreit}}, \bibinfo {author}
  {\bibfnamefont {L.}~\bibnamefont {Jones}}, \bibinfo {author} {\bibfnamefont
  {A.~N.}\ \bibnamefont {Gomez}}, \bibinfo {author} {\bibfnamefont
  {{\L}.}~\bibnamefont {Kaiser}},\ and\ \bibinfo {author} {\bibfnamefont
  {I.}~\bibnamefont {Polosukhin}},\ }\bibfield  {title} {\bibinfo {title}
  {Attention is all you need},\ }in\ \href@noop {} {\emph {\bibinfo {booktitle}
  {Proceedings of the 31st {{International Conference}} on {{Neural Information
  Processing Systems}}}}},\ \bibinfo {series and number} {{{NIPS}}'17}\
  (\bibinfo  {publisher} {Curran Associates Inc.},\ \bibinfo {address} {Red
  Hook, NY, USA},\ \bibinfo {year} {2017})\ pp.\ \bibinfo {pages}
  {6000--6010}\BibitemShut {NoStop}%
\bibitem [{\citenamefont {Pengmei}\ \emph {et~al.}(2023)\citenamefont
  {Pengmei}, \citenamefont {Li}, \citenamefont {chan Tien}, \citenamefont
  {Kondor},\ and\ \citenamefont {Dinner}}]{pengmei2023transformers}%
  \BibitemOpen
  \bibfield  {author} {\bibinfo {author} {\bibfnamefont {Z.}~\bibnamefont
  {Pengmei}}, \bibinfo {author} {\bibfnamefont {Z.}~\bibnamefont {Li}},
  \bibinfo {author} {\bibfnamefont {C.}~\bibnamefont {chan Tien}}, \bibinfo
  {author} {\bibfnamefont {R.}~\bibnamefont {Kondor}},\ and\ \bibinfo {author}
  {\bibfnamefont {A.~R.}\ \bibnamefont {Dinner}},\ }\bibfield  {title}
  {\bibinfo {title} {Transformers are efficient hierarchical chemical graph
  learners},\ }\href@noop {} {\bibfield  {journal} {\bibinfo  {journal} {arXiv
  preprint arXiv:2310.01704}\ } (\bibinfo {year} {2023})}\BibitemShut {NoStop}%
\bibitem [{\citenamefont {Tolstikhin}\ \emph {et~al.}(2021)\citenamefont
  {Tolstikhin}, \citenamefont {Houlsby}, \citenamefont {Kolesnikov},
  \citenamefont {Beyer}, \citenamefont {Zhai}, \citenamefont {Unterthiner},
  \citenamefont {Yung}, \citenamefont {Steiner}, \citenamefont {Keysers},
  \citenamefont {Uszkoreit}, \citenamefont {Lucic},\ and\ \citenamefont
  {Dosovitskiy}}]{tolstikhin2021mlp}%
  \BibitemOpen
  \bibfield  {author} {\bibinfo {author} {\bibfnamefont {I.~O.}\ \bibnamefont
  {Tolstikhin}}, \bibinfo {author} {\bibfnamefont {N.}~\bibnamefont {Houlsby}},
  \bibinfo {author} {\bibfnamefont {A.}~\bibnamefont {Kolesnikov}}, \bibinfo
  {author} {\bibfnamefont {L.}~\bibnamefont {Beyer}}, \bibinfo {author}
  {\bibfnamefont {X.}~\bibnamefont {Zhai}}, \bibinfo {author} {\bibfnamefont
  {T.}~\bibnamefont {Unterthiner}}, \bibinfo {author} {\bibfnamefont
  {J.}~\bibnamefont {Yung}}, \bibinfo {author} {\bibfnamefont {A.}~\bibnamefont
  {Steiner}}, \bibinfo {author} {\bibfnamefont {D.}~\bibnamefont {Keysers}},
  \bibinfo {author} {\bibfnamefont {J.}~\bibnamefont {Uszkoreit}}, \bibinfo
  {author} {\bibfnamefont {M.}~\bibnamefont {Lucic}},\ and\ \bibinfo {author}
  {\bibfnamefont {A.}~\bibnamefont {Dosovitskiy}},\ }\bibfield  {title}
  {\bibinfo {title} {{{MLP-Mixer}}: {{An}} all-{{MLP}} architecture for
  vision},\ }in\ \href@noop {} {\emph {\bibinfo {booktitle} {Advances in
  {{Neural Information Processing Systems}}}}},\ Vol.~\bibinfo {volume} {34}\
  (\bibinfo  {publisher} {Curran Associates, Inc.},\ \bibinfo {year} {2021})\
  pp.\ \bibinfo {pages} {24261--24272}\BibitemShut {NoStop}%
\bibitem [{\citenamefont {Jing}\ \emph {et~al.}(2020)\citenamefont {Jing},
  \citenamefont {Eismann}, \citenamefont {Suriana}, \citenamefont {Townshend},\
  and\ \citenamefont {Dror}}]{jing2020learning}%
  \BibitemOpen
  \bibfield  {author} {\bibinfo {author} {\bibfnamefont {B.}~\bibnamefont
  {Jing}}, \bibinfo {author} {\bibfnamefont {S.}~\bibnamefont {Eismann}},
  \bibinfo {author} {\bibfnamefont {P.}~\bibnamefont {Suriana}}, \bibinfo
  {author} {\bibfnamefont {R.~J.~L.}\ \bibnamefont {Townshend}},\ and\ \bibinfo
  {author} {\bibfnamefont {R.}~\bibnamefont {Dror}},\ }\bibfield  {title}
  {\bibinfo {title} {Learning from protein structure with geometric vector
  perceptrons},\ }in\ \href@noop {} {\emph {\bibinfo {booktitle} {International
  Conference on Learning Representations}}}\ (\bibinfo {year}
  {2020})\BibitemShut {NoStop}%
\bibitem [{\citenamefont {Kipf}\ and\ \citenamefont
  {Welling}(2016)}]{kipf2016semi}%
  \BibitemOpen
  \bibfield  {author} {\bibinfo {author} {\bibfnamefont {T.~N.}\ \bibnamefont
  {Kipf}}\ and\ \bibinfo {author} {\bibfnamefont {M.}~\bibnamefont {Welling}},\
  }\bibfield  {title} {\bibinfo {title} {Semi-supervised classification with
  graph convolutional networks},\ }\href@noop {} {\bibfield  {journal}
  {\bibinfo  {journal} {arXiv preprint arXiv:1609.02907}\ } (\bibinfo {year}
  {2016})}\BibitemShut {NoStop}%
\bibitem [{\citenamefont {Morris}\ \emph {et~al.}(2019)\citenamefont {Morris},
  \citenamefont {Ritzert}, \citenamefont {Fey}, \citenamefont {Hamilton},
  \citenamefont {Lenssen}, \citenamefont {Rattan},\ and\ \citenamefont
  {Grohe}}]{morris2019weisfeiler}%
  \BibitemOpen
  \bibfield  {author} {\bibinfo {author} {\bibfnamefont {C.}~\bibnamefont
  {Morris}}, \bibinfo {author} {\bibfnamefont {M.}~\bibnamefont {Ritzert}},
  \bibinfo {author} {\bibfnamefont {M.}~\bibnamefont {Fey}}, \bibinfo {author}
  {\bibfnamefont {W.~L.}\ \bibnamefont {Hamilton}}, \bibinfo {author}
  {\bibfnamefont {J.~E.}\ \bibnamefont {Lenssen}}, \bibinfo {author}
  {\bibfnamefont {G.}~\bibnamefont {Rattan}},\ and\ \bibinfo {author}
  {\bibfnamefont {M.}~\bibnamefont {Grohe}},\ }\bibfield  {title} {\bibinfo
  {title} {{Weisfeiler} and {Leman} go neural: Higher-order graph neural
  networks},\ }in\ \href@noop {} {\emph {\bibinfo {booktitle} {Proceedings of
  the AAAI conference on artificial intelligence}}},\ Vol.~\bibinfo {volume}
  {33}\ (\bibinfo {year} {2019})\ pp.\ \bibinfo {pages}
  {4602--4609}\BibitemShut {NoStop}%
\bibitem [{\citenamefont {Bresson}\ and\ \citenamefont
  {Laurent}(2017)}]{bresson2017residual}%
  \BibitemOpen
  \bibfield  {author} {\bibinfo {author} {\bibfnamefont {X.}~\bibnamefont
  {Bresson}}\ and\ \bibinfo {author} {\bibfnamefont {T.}~\bibnamefont
  {Laurent}},\ }\bibfield  {title} {\bibinfo {title} {Residual gated graph
  {ConvNets}},\ }\href@noop {} {\bibfield  {journal} {\bibinfo  {journal}
  {arXiv preprint arXiv:1711.07553}\ } (\bibinfo {year} {2017})}\BibitemShut
  {NoStop}%
\bibitem [{\citenamefont {Du}\ \emph {et~al.}(2017)\citenamefont {Du},
  \citenamefont {Zhang}, \citenamefont {Wu}, \citenamefont {Moura},\ and\
  \citenamefont {Kar}}]{du2017topology}%
  \BibitemOpen
  \bibfield  {author} {\bibinfo {author} {\bibfnamefont {J.}~\bibnamefont
  {Du}}, \bibinfo {author} {\bibfnamefont {S.}~\bibnamefont {Zhang}}, \bibinfo
  {author} {\bibfnamefont {G.}~\bibnamefont {Wu}}, \bibinfo {author}
  {\bibfnamefont {J.~M.}\ \bibnamefont {Moura}},\ and\ \bibinfo {author}
  {\bibfnamefont {S.}~\bibnamefont {Kar}},\ }\bibfield  {title} {\bibinfo
  {title} {Topology adaptive graph convolutional networks},\ }\href@noop {}
  {\bibfield  {journal} {\bibinfo  {journal} {arXiv preprint arXiv:1710.10370}\
  } (\bibinfo {year} {2017})}\BibitemShut {NoStop}%
\bibitem [{\citenamefont {Fey}\ and\ \citenamefont
  {Lenssen}(2019)}]{Fey/Lenssen/2019}%
  \BibitemOpen
  \bibfield  {author} {\bibinfo {author} {\bibfnamefont {M.}~\bibnamefont
  {Fey}}\ and\ \bibinfo {author} {\bibfnamefont {J.~E.}\ \bibnamefont
  {Lenssen}},\ }\bibfield  {title} {\bibinfo {title} {Fast graph representation
  learning with {PyTorch Geometric}},\ }in\ \href@noop {} {\emph {\bibinfo
  {booktitle} {ICLR Workshop on Representation Learning on Graphs and
  Manifolds}}}\ (\bibinfo {year} {2019})\BibitemShut {NoStop}%
\bibitem [{\citenamefont {Chen}\ \emph {et~al.}(2019)\citenamefont {Chen},
  \citenamefont {Sidky},\ and\ \citenamefont {Ferguson}}]{chen_nonlinear_2019}%
  \BibitemOpen
  \bibfield  {author} {\bibinfo {author} {\bibfnamefont {W.}~\bibnamefont
  {Chen}}, \bibinfo {author} {\bibfnamefont {H.}~\bibnamefont {Sidky}},\ and\
  \bibinfo {author} {\bibfnamefont {A.~L.}\ \bibnamefont {Ferguson}},\
  }\bibfield  {title} {\bibinfo {title} {Nonlinear discovery of slow molecular
  modes using state-free reversible {VAMPnets}},\ }\href
  {https://doi.org/10.1063/1.5092521} {\bibfield  {journal} {\bibinfo
  {journal} {Journal of Chemical Physics}\ }\textbf {\bibinfo {volume} {150}},\
  \bibinfo {pages} {214114} (\bibinfo {year} {2019})}\BibitemShut {NoStop}%
\bibitem [{\citenamefont {Wu}\ and\ \citenamefont
  {No{\'e}}(2020)}]{wu_variational_2020}%
  \BibitemOpen
  \bibfield  {author} {\bibinfo {author} {\bibfnamefont {H.}~\bibnamefont
  {Wu}}\ and\ \bibinfo {author} {\bibfnamefont {F.}~\bibnamefont {No{\'e}}},\
  }\bibfield  {title} {\bibinfo {title} {Variational approach for learning
  {{Markov}} processes from time series data},\ }\href
  {https://doi.org/10.1007/s00332-019-09567-y} {\bibfield  {journal} {\bibinfo
  {journal} {Journal of Nonlinear Science}\ }\textbf {\bibinfo {volume} {30}},\
  \bibinfo {pages} {23} (\bibinfo {year} {2020})}\BibitemShut {NoStop}%
\bibitem [{\citenamefont {Lorpaiboon}\ \emph {et~al.}(2020)\citenamefont
  {Lorpaiboon}, \citenamefont {Thiede}, \citenamefont {Webber}, \citenamefont
  {Weare},\ and\ \citenamefont {Dinner}}]{lorpaiboon_integrated_2020}%
  \BibitemOpen
  \bibfield  {author} {\bibinfo {author} {\bibfnamefont {C.}~\bibnamefont
  {Lorpaiboon}}, \bibinfo {author} {\bibfnamefont {E.~H.}\ \bibnamefont
  {Thiede}}, \bibinfo {author} {\bibfnamefont {R.~J.}\ \bibnamefont {Webber}},
  \bibinfo {author} {\bibfnamefont {J.}~\bibnamefont {Weare}},\ and\ \bibinfo
  {author} {\bibfnamefont {A.~R.}\ \bibnamefont {Dinner}},\ }\bibfield  {title}
  {\bibinfo {title} {Integrated variational approach to conformational
  dynamics: A robust strategy for identifying eigenfunctions of dynamical
  operators},\ }\href {https://doi.org/10.1021/acs.jpcb.0c06477} {\bibfield
  {journal} {\bibinfo  {journal} {Journal of Physical Chemistry B}\ }\textbf
  {\bibinfo {volume} {124}},\ \bibinfo {pages} {9354} (\bibinfo {year}
  {2020})}\BibitemShut {NoStop}%
\bibitem [{\citenamefont {Strahan}\ \emph
  {et~al.}(2023{\natexlab{a}})\citenamefont {Strahan}, \citenamefont {Finkel},
  \citenamefont {Dinner},\ and\ \citenamefont
  {Weare}}]{strahan_predicting_2023}%
  \BibitemOpen
  \bibfield  {author} {\bibinfo {author} {\bibfnamefont {J.}~\bibnamefont
  {Strahan}}, \bibinfo {author} {\bibfnamefont {J.}~\bibnamefont {Finkel}},
  \bibinfo {author} {\bibfnamefont {A.~R.}\ \bibnamefont {Dinner}},\ and\
  \bibinfo {author} {\bibfnamefont {J.}~\bibnamefont {Weare}},\ }\bibfield
  {title} {\bibinfo {title} {Predicting rare events using neural networks and
  short-trajectory data},\ }\href@noop {} {\bibfield  {journal} {\bibinfo
  {journal} {Journal of computational physics}\ }\textbf {\bibinfo {volume}
  {488}},\ \bibinfo {pages} {112152} (\bibinfo {year}
  {2023}{\natexlab{a}})}\BibitemShut {NoStop}%
\bibitem [{\citenamefont {Strahan}\ \emph
  {et~al.}(2023{\natexlab{b}})\citenamefont {Strahan}, \citenamefont {Guo},
  \citenamefont {Lorpaiboon}, \citenamefont {Dinner},\ and\ \citenamefont
  {Weare}}]{strahan_inexact_2023}%
  \BibitemOpen
  \bibfield  {author} {\bibinfo {author} {\bibfnamefont {J.}~\bibnamefont
  {Strahan}}, \bibinfo {author} {\bibfnamefont {S.~C.}\ \bibnamefont {Guo}},
  \bibinfo {author} {\bibfnamefont {C.}~\bibnamefont {Lorpaiboon}}, \bibinfo
  {author} {\bibfnamefont {A.~R.}\ \bibnamefont {Dinner}},\ and\ \bibinfo
  {author} {\bibfnamefont {J.}~\bibnamefont {Weare}},\ }\bibfield  {title}
  {\bibinfo {title} {Inexact iterative numerical linear algebra for neural
  network-based spectral estimation and rare-event prediction},\ }\href
  {https://doi.org/10.1063/5.0151309} {\bibfield  {journal} {\bibinfo
  {journal} {Journal of Chemical Physics}\ }\textbf {\bibinfo {volume} {159}},\
  \bibinfo {pages} {014110} (\bibinfo {year} {2023}{\natexlab{b}})}\BibitemShut
  {NoStop}%
\bibitem [{\citenamefont {Tomczak}\ and\ \citenamefont
  {Welling}(2018)}]{tomczak2018vae}%
  \BibitemOpen
  \bibfield  {author} {\bibinfo {author} {\bibfnamefont {J.}~\bibnamefont
  {Tomczak}}\ and\ \bibinfo {author} {\bibfnamefont {M.}~\bibnamefont
  {Welling}},\ }\bibfield  {title} {\bibinfo {title} {{{VAE}} with a
  {{VampPrior}}},\ }in\ \href@noop {} {\emph {\bibinfo {booktitle} {Proceedings
  of the {{Twenty-First International Conference}} on {{Artificial
  Intelligence}} and {{Statistics}}}}}\ (\bibinfo  {publisher} {PMLR},\
  \bibinfo {year} {2018})\ pp.\ \bibinfo {pages} {1214--1223}\BibitemShut
  {NoStop}%
\bibitem [{\citenamefont {M{\"u}ller}\ and\ \citenamefont
  {Schulz}(1992)}]{muller_structure_1992}%
  \BibitemOpen
  \bibfield  {author} {\bibinfo {author} {\bibfnamefont {C.~W.}\ \bibnamefont
  {M{\"u}ller}}\ and\ \bibinfo {author} {\bibfnamefont {G.~E.}\ \bibnamefont
  {Schulz}},\ }\bibfield  {title} {\bibinfo {title} {Structure of the complex
  between adenylate kinase from {{{\emph{Escherichia}}}}{\emph{ coli}} and the
  inhibitor {{Ap5A}} refined at 1.9 {{{\AA}}} resolution},\ }\href@noop {}
  {\bibfield  {journal} {\bibinfo  {journal} {Journal of Molecular Biology}\
  }\textbf {\bibinfo {volume} {224}},\ \bibinfo {pages} {159} (\bibinfo {year}
  {1992})}\BibitemShut {NoStop}%
\bibitem [{\citenamefont {Beckstein}\ \emph {et~al.}(2009)\citenamefont
  {Beckstein}, \citenamefont {Denning}, \citenamefont {Perilla},\ and\
  \citenamefont {Woolf}}]{beckstein_zipping_2009}%
  \BibitemOpen
  \bibfield  {author} {\bibinfo {author} {\bibfnamefont {O.}~\bibnamefont
  {Beckstein}}, \bibinfo {author} {\bibfnamefont {E.~J.}\ \bibnamefont
  {Denning}}, \bibinfo {author} {\bibfnamefont {J.~R.}\ \bibnamefont
  {Perilla}},\ and\ \bibinfo {author} {\bibfnamefont {T.~B.}\ \bibnamefont
  {Woolf}},\ }\bibfield  {title} {\bibinfo {title} {Zipping and unzipping of
  adenylate kinase: {{Atomistic}} insights into the ensemble of open
  {$\leftrightarrow$} closed transitions},\ }\href@noop {} {\bibfield
  {journal} {\bibinfo  {journal} {Journal of Molecular Biology}\ }\textbf
  {\bibinfo {volume} {394}},\ \bibinfo {pages} {160} (\bibinfo {year}
  {2009})}\BibitemShut {NoStop}%
\bibitem [{\citenamefont {Seyler}\ \emph {et~al.}(2015)\citenamefont {Seyler},
  \citenamefont {Kumar}, \citenamefont {Thorpe},\ and\ \citenamefont
  {Beckstein}}]{seyler_path_2015}%
  \BibitemOpen
  \bibfield  {author} {\bibinfo {author} {\bibfnamefont {S.~L.}\ \bibnamefont
  {Seyler}}, \bibinfo {author} {\bibfnamefont {A.}~\bibnamefont {Kumar}},
  \bibinfo {author} {\bibfnamefont {M.~F.}\ \bibnamefont {Thorpe}},\ and\
  \bibinfo {author} {\bibfnamefont {O.}~\bibnamefont {Beckstein}},\ }\bibfield
  {title} {\bibinfo {title} {Path similarity analysis: a method for quantifying
  macromolecular pathways},\ }\href@noop {} {\bibfield  {journal} {\bibinfo
  {journal} {PLOS Computational Biology}\ }\textbf {\bibinfo {volume} {11}},\
  \bibinfo {pages} {e1004568} (\bibinfo {year} {2015})}\BibitemShut {NoStop}%
\bibitem [{\citenamefont {Zheng}\ and\ \citenamefont
  {Cui}(2018)}]{zheng_multiple_2018a}%
  \BibitemOpen
  \bibfield  {author} {\bibinfo {author} {\bibfnamefont {Y.}~\bibnamefont
  {Zheng}}\ and\ \bibinfo {author} {\bibfnamefont {Q.}~\bibnamefont {Cui}},\
  }\bibfield  {title} {\bibinfo {title} {Multiple pathways and time scales for
  conformational transitions in apo-adenylate kinase},\ }\href@noop {}
  {\bibfield  {journal} {\bibinfo  {journal} {Journal of Chemical Theory and
  Computation}\ }\textbf {\bibinfo {volume} {14}},\ \bibinfo {pages} {1716}
  (\bibinfo {year} {2018})}\BibitemShut {NoStop}%
\bibitem [{\citenamefont {Stiller}\ \emph {et~al.}(2019)\citenamefont
  {Stiller}, \citenamefont {Jordan~Kerns}, \citenamefont {Hoemberger},
  \citenamefont {Cho}, \citenamefont {Otten}, \citenamefont {Hagan},\ and\
  \citenamefont {Kern}}]{stiller_probing_2019}%
  \BibitemOpen
  \bibfield  {author} {\bibinfo {author} {\bibfnamefont {J.~B.}\ \bibnamefont
  {Stiller}}, \bibinfo {author} {\bibfnamefont {S.}~\bibnamefont
  {Jordan~Kerns}}, \bibinfo {author} {\bibfnamefont {M.}~\bibnamefont
  {Hoemberger}}, \bibinfo {author} {\bibfnamefont {Y.-J.}\ \bibnamefont {Cho}},
  \bibinfo {author} {\bibfnamefont {R.}~\bibnamefont {Otten}}, \bibinfo
  {author} {\bibfnamefont {M.~F.}\ \bibnamefont {Hagan}},\ and\ \bibinfo
  {author} {\bibfnamefont {D.}~\bibnamefont {Kern}},\ }\bibfield  {title}
  {\bibinfo {title} {Probing the transition state in enzyme catalysis by
  high-pressure {{NMR}} dynamics},\ }\href@noop {} {\bibfield  {journal}
  {\bibinfo  {journal} {Nature Catalysis}\ }\textbf {\bibinfo {volume} {2}},\
  \bibinfo {pages} {726} (\bibinfo {year} {2019})}\BibitemShut {NoStop}%
\bibitem [{\citenamefont {Stiller}\ \emph {et~al.}(2022)\citenamefont
  {Stiller}, \citenamefont {Otten}, \citenamefont {H{\"a}ussinger},
  \citenamefont {Rieder}, \citenamefont {Theobald},\ and\ \citenamefont
  {Kern}}]{stiller_structure_2022}%
  \BibitemOpen
  \bibfield  {author} {\bibinfo {author} {\bibfnamefont {J.~B.}\ \bibnamefont
  {Stiller}}, \bibinfo {author} {\bibfnamefont {R.}~\bibnamefont {Otten}},
  \bibinfo {author} {\bibfnamefont {D.}~\bibnamefont {H{\"a}ussinger}},
  \bibinfo {author} {\bibfnamefont {P.~S.}\ \bibnamefont {Rieder}}, \bibinfo
  {author} {\bibfnamefont {D.~L.}\ \bibnamefont {Theobald}},\ and\ \bibinfo
  {author} {\bibfnamefont {D.}~\bibnamefont {Kern}},\ }\bibfield  {title}
  {\bibinfo {title} {Structure determination of high-energy states in a dynamic
  protein ensemble},\ }\href@noop {} {\bibfield  {journal} {\bibinfo  {journal}
  {Nature}\ ,\ \bibinfo {pages} {1}} (\bibinfo {year} {2022})}\BibitemShut
  {NoStop}%
\bibitem [{\citenamefont {Li}\ and\ \citenamefont
  {Bruschweiler}(2011)}]{li2011iterative}%
  \BibitemOpen
  \bibfield  {author} {\bibinfo {author} {\bibfnamefont {D.-W.}\ \bibnamefont
  {Li}}\ and\ \bibinfo {author} {\bibfnamefont {R.}~\bibnamefont
  {Bruschweiler}},\ }\bibfield  {title} {\bibinfo {title} {Iterative
  optimization of molecular mechanics force fields from nmr data of full-length
  proteins},\ }\href@noop {} {\bibfield  {journal} {\bibinfo  {journal}
  {Journal of chemical theory and computation}\ }\textbf {\bibinfo {volume}
  {7}},\ \bibinfo {pages} {1773} (\bibinfo {year} {2011})}\BibitemShut
  {NoStop}%
\bibitem [{\citenamefont {Chen}\ \emph {et~al.}(2020)\citenamefont {Chen},
  \citenamefont {Malone}, \citenamefont {Llewellyn}, \citenamefont {Grasso},
  \citenamefont {Shelton}, \citenamefont {Olinares}, \citenamefont {Maruthi},
  \citenamefont {Eng}, \citenamefont {Vatandaslar}, \citenamefont {Chait},
  \citenamefont {Kapoor}, \citenamefont {Darst},\ and\ \citenamefont
  {Campbell}}]{chen_structural_2020}%
  \BibitemOpen
  \bibfield  {author} {\bibinfo {author} {\bibfnamefont {J.}~\bibnamefont
  {Chen}}, \bibinfo {author} {\bibfnamefont {B.}~\bibnamefont {Malone}},
  \bibinfo {author} {\bibfnamefont {E.}~\bibnamefont {Llewellyn}}, \bibinfo
  {author} {\bibfnamefont {M.}~\bibnamefont {Grasso}}, \bibinfo {author}
  {\bibfnamefont {P.~M.~M.}\ \bibnamefont {Shelton}}, \bibinfo {author}
  {\bibfnamefont {P.~D.~B.}\ \bibnamefont {Olinares}}, \bibinfo {author}
  {\bibfnamefont {K.}~\bibnamefont {Maruthi}}, \bibinfo {author} {\bibfnamefont
  {E.~T.}\ \bibnamefont {Eng}}, \bibinfo {author} {\bibfnamefont
  {H.}~\bibnamefont {Vatandaslar}}, \bibinfo {author} {\bibfnamefont {B.~T.}\
  \bibnamefont {Chait}}, \bibinfo {author} {\bibfnamefont {T.~M.}\ \bibnamefont
  {Kapoor}}, \bibinfo {author} {\bibfnamefont {S.~A.}\ \bibnamefont {Darst}},\
  and\ \bibinfo {author} {\bibfnamefont {E.~A.}\ \bibnamefont {Campbell}},\
  }\bibfield  {title} {\bibinfo {title} {Structural basis for
  helicase-polymerase coupling in the {{SARS-CoV-2}} replication-transcription
  complex},\ }\href@noop {} {\bibfield  {journal} {\bibinfo  {journal} {Cell}\
  }\textbf {\bibinfo {volume} {182}},\ \bibinfo {pages} {1560} (\bibinfo {year}
  {2020})}\BibitemShut {NoStop}%
\bibitem [{\citenamefont {Yan}\ \emph {et~al.}(2020)\citenamefont {Yan},
  \citenamefont {Zhang}, \citenamefont {Ge}, \citenamefont {Zheng},
  \citenamefont {Gao}, \citenamefont {Wang}, \citenamefont {Jia}, \citenamefont
  {Wang}, \citenamefont {Huang}, \citenamefont {Li}, \citenamefont {Wang},
  \citenamefont {Rao},\ and\ \citenamefont {Lou}}]{yan_architecture_2020}%
  \BibitemOpen
  \bibfield  {author} {\bibinfo {author} {\bibfnamefont {L.}~\bibnamefont
  {Yan}}, \bibinfo {author} {\bibfnamefont {Y.}~\bibnamefont {Zhang}}, \bibinfo
  {author} {\bibfnamefont {J.}~\bibnamefont {Ge}}, \bibinfo {author}
  {\bibfnamefont {L.}~\bibnamefont {Zheng}}, \bibinfo {author} {\bibfnamefont
  {Y.}~\bibnamefont {Gao}}, \bibinfo {author} {\bibfnamefont {T.}~\bibnamefont
  {Wang}}, \bibinfo {author} {\bibfnamefont {Z.}~\bibnamefont {Jia}}, \bibinfo
  {author} {\bibfnamefont {H.}~\bibnamefont {Wang}}, \bibinfo {author}
  {\bibfnamefont {Y.}~\bibnamefont {Huang}}, \bibinfo {author} {\bibfnamefont
  {M.}~\bibnamefont {Li}}, \bibinfo {author} {\bibfnamefont {Q.}~\bibnamefont
  {Wang}}, \bibinfo {author} {\bibfnamefont {Z.}~\bibnamefont {Rao}},\ and\
  \bibinfo {author} {\bibfnamefont {Z.}~\bibnamefont {Lou}},\ }\bibfield
  {title} {\bibinfo {title} {Architecture of a {{SARS-CoV-2}} mini replication
  and transcription complex},\ }\href@noop {} {\bibfield  {journal} {\bibinfo
  {journal} {Nature Communications}\ }\textbf {\bibinfo {volume} {11}},\
  \bibinfo {pages} {5874} (\bibinfo {year} {2020})}\BibitemShut {NoStop}%
\bibitem [{\citenamefont {Yan}\ \emph {et~al.}(2021)\citenamefont {Yan},
  \citenamefont {Ge}, \citenamefont {Zheng}, \citenamefont {Zhang},
  \citenamefont {Gao}, \citenamefont {Wang}, \citenamefont {Huang},
  \citenamefont {Yang}, \citenamefont {Gao}, \citenamefont {Li}, \citenamefont
  {Liu}, \citenamefont {Wang}, \citenamefont {Li}, \citenamefont {Chen},
  \citenamefont {Guddat}, \citenamefont {Wang}, \citenamefont {Rao},\ and\
  \citenamefont {Lou}}]{yan_cryoem_2021}%
  \BibitemOpen
  \bibfield  {author} {\bibinfo {author} {\bibfnamefont {L.}~\bibnamefont
  {Yan}}, \bibinfo {author} {\bibfnamefont {J.}~\bibnamefont {Ge}}, \bibinfo
  {author} {\bibfnamefont {L.}~\bibnamefont {Zheng}}, \bibinfo {author}
  {\bibfnamefont {Y.}~\bibnamefont {Zhang}}, \bibinfo {author} {\bibfnamefont
  {Y.}~\bibnamefont {Gao}}, \bibinfo {author} {\bibfnamefont {T.}~\bibnamefont
  {Wang}}, \bibinfo {author} {\bibfnamefont {Y.}~\bibnamefont {Huang}},
  \bibinfo {author} {\bibfnamefont {Y.}~\bibnamefont {Yang}}, \bibinfo {author}
  {\bibfnamefont {S.}~\bibnamefont {Gao}}, \bibinfo {author} {\bibfnamefont
  {M.}~\bibnamefont {Li}}, \bibinfo {author} {\bibfnamefont {Z.}~\bibnamefont
  {Liu}}, \bibinfo {author} {\bibfnamefont {H.}~\bibnamefont {Wang}}, \bibinfo
  {author} {\bibfnamefont {Y.}~\bibnamefont {Li}}, \bibinfo {author}
  {\bibfnamefont {Y.}~\bibnamefont {Chen}}, \bibinfo {author} {\bibfnamefont
  {L.~W.}\ \bibnamefont {Guddat}}, \bibinfo {author} {\bibfnamefont
  {Q.}~\bibnamefont {Wang}}, \bibinfo {author} {\bibfnamefont {Z.}~\bibnamefont
  {Rao}},\ and\ \bibinfo {author} {\bibfnamefont {Z.}~\bibnamefont {Lou}},\
  }\bibfield  {title} {\bibinfo {title} {Cryo-{{EM}} structure of an extended
  {{SARS-CoV-2}} replication and transcription complex reveals an intermediate
  state in cap synthesis},\ }\href@noop {} {\bibfield  {journal} {\bibinfo
  {journal} {Cell}\ }\textbf {\bibinfo {volume} {184}},\ \bibinfo {pages} {184}
  (\bibinfo {year} {2021})}\BibitemShut {NoStop}%
\bibitem [{\citenamefont {Pengmei}\ \emph
  {et~al.}(2024{\natexlab{c}})\citenamefont {Pengmei}, \citenamefont {Liu},\
  and\ \citenamefont {Shu}}]{pengmei2024beyond}%
  \BibitemOpen
  \bibfield  {author} {\bibinfo {author} {\bibfnamefont {Z.}~\bibnamefont
  {Pengmei}}, \bibinfo {author} {\bibfnamefont {J.}~\bibnamefont {Liu}},\ and\
  \bibinfo {author} {\bibfnamefont {Y.}~\bibnamefont {Shu}},\ }\bibfield
  {title} {\bibinfo {title} {Beyond {MD17}: the reactive {xxMD} dataset},\
  }\href@noop {} {\bibfield  {journal} {\bibinfo  {journal} {Scientific Data}\
  }\textbf {\bibinfo {volume} {11}},\ \bibinfo {pages} {222} (\bibinfo {year}
  {2024}{\natexlab{c}})}\BibitemShut {NoStop}%
\bibitem [{\citenamefont {Ying}\ \emph {et~al.}(2021)\citenamefont {Ying},
  \citenamefont {Cai}, \citenamefont {Luo}, \citenamefont {Zheng},
  \citenamefont {Ke}, \citenamefont {He}, \citenamefont {Shen},\ and\
  \citenamefont {Liu}}]{ying2021transformers}%
  \BibitemOpen
  \bibfield  {author} {\bibinfo {author} {\bibfnamefont {C.}~\bibnamefont
  {Ying}}, \bibinfo {author} {\bibfnamefont {T.}~\bibnamefont {Cai}}, \bibinfo
  {author} {\bibfnamefont {S.}~\bibnamefont {Luo}}, \bibinfo {author}
  {\bibfnamefont {S.}~\bibnamefont {Zheng}}, \bibinfo {author} {\bibfnamefont
  {G.}~\bibnamefont {Ke}}, \bibinfo {author} {\bibfnamefont {D.}~\bibnamefont
  {He}}, \bibinfo {author} {\bibfnamefont {Y.}~\bibnamefont {Shen}},\ and\
  \bibinfo {author} {\bibfnamefont {T.-Y.}\ \bibnamefont {Liu}},\ }\bibfield
  {title} {\bibinfo {title} {Do transformers really perform badly for graph
  representation?},\ }\href@noop {} {\bibfield  {journal} {\bibinfo  {journal}
  {Advances in neural information processing systems}\ }\textbf {\bibinfo
  {volume} {34}},\ \bibinfo {pages} {28877} (\bibinfo {year}
  {2021})}\BibitemShut {NoStop}%
\bibitem [{\citenamefont {Shi}\ \emph {et~al.}(2022)\citenamefont {Shi},
  \citenamefont {Zheng}, \citenamefont {Ke}, \citenamefont {Shen},
  \citenamefont {You}, \citenamefont {He}, \citenamefont {Luo}, \citenamefont
  {Liu}, \citenamefont {He},\ and\ \citenamefont {Liu}}]{shi2022benchmarking}%
  \BibitemOpen
  \bibfield  {author} {\bibinfo {author} {\bibfnamefont {Y.}~\bibnamefont
  {Shi}}, \bibinfo {author} {\bibfnamefont {S.}~\bibnamefont {Zheng}}, \bibinfo
  {author} {\bibfnamefont {G.}~\bibnamefont {Ke}}, \bibinfo {author}
  {\bibfnamefont {Y.}~\bibnamefont {Shen}}, \bibinfo {author} {\bibfnamefont
  {J.}~\bibnamefont {You}}, \bibinfo {author} {\bibfnamefont {J.}~\bibnamefont
  {He}}, \bibinfo {author} {\bibfnamefont {S.}~\bibnamefont {Luo}}, \bibinfo
  {author} {\bibfnamefont {C.}~\bibnamefont {Liu}}, \bibinfo {author}
  {\bibfnamefont {D.}~\bibnamefont {He}},\ and\ \bibinfo {author}
  {\bibfnamefont {T.-Y.}\ \bibnamefont {Liu}},\ }\bibfield  {title} {\bibinfo
  {title} {Benchmarking graphormer on large-scale molecular modeling
  datasets},\ }\href@noop {} {\bibfield  {journal} {\bibinfo  {journal} {arXiv
  preprint arXiv:2203.04810}\ } (\bibinfo {year} {2022})}\BibitemShut {NoStop}%
\bibitem [{\citenamefont {Zheng}\ \emph {et~al.}(2024)\citenamefont {Zheng},
  \citenamefont {He}, \citenamefont {Liu}, \citenamefont {Shi}, \citenamefont
  {Lu}, \citenamefont {Feng}, \citenamefont {Ju}, \citenamefont {Wang},
  \citenamefont {Zhu}, \citenamefont {Min} \emph
  {et~al.}}]{zheng2024predicting}%
  \BibitemOpen
  \bibfield  {author} {\bibinfo {author} {\bibfnamefont {S.}~\bibnamefont
  {Zheng}}, \bibinfo {author} {\bibfnamefont {J.}~\bibnamefont {He}}, \bibinfo
  {author} {\bibfnamefont {C.}~\bibnamefont {Liu}}, \bibinfo {author}
  {\bibfnamefont {Y.}~\bibnamefont {Shi}}, \bibinfo {author} {\bibfnamefont
  {Z.}~\bibnamefont {Lu}}, \bibinfo {author} {\bibfnamefont {W.}~\bibnamefont
  {Feng}}, \bibinfo {author} {\bibfnamefont {F.}~\bibnamefont {Ju}}, \bibinfo
  {author} {\bibfnamefont {J.}~\bibnamefont {Wang}}, \bibinfo {author}
  {\bibfnamefont {J.}~\bibnamefont {Zhu}}, \bibinfo {author} {\bibfnamefont
  {Y.}~\bibnamefont {Min}}, \emph {et~al.},\ }\bibfield  {title} {\bibinfo
  {title} {Predicting equilibrium distributions for molecular systems with deep
  learning},\ }\href@noop {} {\bibfield  {journal} {\bibinfo  {journal} {Nature
  Machine Intelligence}\ }\textbf {\bibinfo {volume} {6}},\ \bibinfo {pages}
  {558} (\bibinfo {year} {2024})}\BibitemShut {NoStop}%
\bibitem [{\citenamefont {Kosmala}\ \emph {et~al.}(2023)\citenamefont
  {Kosmala}, \citenamefont {Gasteiger}, \citenamefont {Gao},\ and\
  \citenamefont {G{\"u}nnemann}}]{kosmala2023ewald}%
  \BibitemOpen
  \bibfield  {author} {\bibinfo {author} {\bibfnamefont {A.}~\bibnamefont
  {Kosmala}}, \bibinfo {author} {\bibfnamefont {J.}~\bibnamefont {Gasteiger}},
  \bibinfo {author} {\bibfnamefont {N.}~\bibnamefont {Gao}},\ and\ \bibinfo
  {author} {\bibfnamefont {S.}~\bibnamefont {G{\"u}nnemann}},\ }\bibfield
  {title} {\bibinfo {title} {Ewald-based long-range message passing for
  molecular graphs},\ }in\ \href@noop {} {\emph {\bibinfo {booktitle}
  {International Conference on Machine Learning}}}\ (\bibinfo {organization}
  {PMLR},\ \bibinfo {year} {2023})\ pp.\ \bibinfo {pages}
  {17544--17563}\BibitemShut {NoStop}%
\bibitem [{\citenamefont {Wang}\ \emph
  {et~al.}(2024{\natexlab{b}})\citenamefont {Wang}, \citenamefont {Cheng},
  \citenamefont {Li}, \citenamefont {Ren}, \citenamefont {Shao}, \citenamefont
  {Liu}, \citenamefont {Heng},\ and\ \citenamefont {Zheng}}]{wang2024neural}%
  \BibitemOpen
  \bibfield  {author} {\bibinfo {author} {\bibfnamefont {Y.}~\bibnamefont
  {Wang}}, \bibinfo {author} {\bibfnamefont {C.}~\bibnamefont {Cheng}},
  \bibinfo {author} {\bibfnamefont {S.}~\bibnamefont {Li}}, \bibinfo {author}
  {\bibfnamefont {Y.}~\bibnamefont {Ren}}, \bibinfo {author} {\bibfnamefont
  {B.}~\bibnamefont {Shao}}, \bibinfo {author} {\bibfnamefont {G.}~\bibnamefont
  {Liu}}, \bibinfo {author} {\bibfnamefont {P.-A.}\ \bibnamefont {Heng}},\ and\
  \bibinfo {author} {\bibfnamefont {N.}~\bibnamefont {Zheng}},\ }\bibfield
  {title} {\bibinfo {title} {Neural {P}$^3${M}: A long-range interaction
  modeling enhancer for geometric {G}{N}{N}s},\ }\href@noop {} {\bibfield
  {journal} {\bibinfo  {journal} {Advances in Neural Information Processing
  Systems}\ }\textbf {\bibinfo {volume} {37}},\ \bibinfo {pages} {120336}
  (\bibinfo {year} {2024}{\natexlab{b}})}\BibitemShut {NoStop}%
\bibitem [{\citenamefont {Caruso}\ \emph {et~al.}(2025)\citenamefont {Caruso},
  \citenamefont {Venturin}, \citenamefont {Giambagli}, \citenamefont {Rolando},
  \citenamefont {No{\'e}},\ and\ \citenamefont
  {Clementi}}]{caruso2025extending}%
  \BibitemOpen
  \bibfield  {author} {\bibinfo {author} {\bibfnamefont {A.}~\bibnamefont
  {Caruso}}, \bibinfo {author} {\bibfnamefont {J.}~\bibnamefont {Venturin}},
  \bibinfo {author} {\bibfnamefont {L.}~\bibnamefont {Giambagli}}, \bibinfo
  {author} {\bibfnamefont {E.}~\bibnamefont {Rolando}}, \bibinfo {author}
  {\bibfnamefont {F.}~\bibnamefont {No{\'e}}},\ and\ \bibinfo {author}
  {\bibfnamefont {C.}~\bibnamefont {Clementi}},\ }\bibfield  {title} {\bibinfo
  {title} {Extending the {R}{A}{N}{G}{E} of graph neural networks: Relaying
  attention nodes for global encoding},\ }\href@noop {} {\bibfield  {journal}
  {\bibinfo  {journal} {arXiv preprint arXiv:2502.13797}\ } (\bibinfo {year}
  {2025})}\BibitemShut {NoStop}%
\bibitem [{\citenamefont {Fey}\ \emph {et~al.}(2020)\citenamefont {Fey},
  \citenamefont {Yuen},\ and\ \citenamefont {Weichert}}]{Fey/etal/2020}%
  \BibitemOpen
  \bibfield  {author} {\bibinfo {author} {\bibfnamefont {M.}~\bibnamefont
  {Fey}}, \bibinfo {author} {\bibfnamefont {J.~G.}\ \bibnamefont {Yuen}},\ and\
  \bibinfo {author} {\bibfnamefont {F.}~\bibnamefont {Weichert}},\ }\bibfield
  {title} {\bibinfo {title} {Hierarchical inter-message passing for learning on
  molecular graphs},\ }in\ \href@noop {} {\emph {\bibinfo {booktitle} {ICML
  Graph Representation Learning and Beyond (GRL+) Workhop}}}\ (\bibinfo {year}
  {2020})\BibitemShut {NoStop}%
\bibitem [{\citenamefont {Li}\ \emph {et~al.}(2023)\citenamefont {Li},
  \citenamefont {Wang}, \citenamefont {Huang}, \citenamefont {Yang},
  \citenamefont {Wei}, \citenamefont {Zhang}, \citenamefont {Wang},
  \citenamefont {Wang}, \citenamefont {Shao},\ and\ \citenamefont
  {Liu}}]{li2023long}%
  \BibitemOpen
  \bibfield  {author} {\bibinfo {author} {\bibfnamefont {Y.}~\bibnamefont
  {Li}}, \bibinfo {author} {\bibfnamefont {Y.}~\bibnamefont {Wang}}, \bibinfo
  {author} {\bibfnamefont {L.}~\bibnamefont {Huang}}, \bibinfo {author}
  {\bibfnamefont {H.}~\bibnamefont {Yang}}, \bibinfo {author} {\bibfnamefont
  {X.}~\bibnamefont {Wei}}, \bibinfo {author} {\bibfnamefont {J.}~\bibnamefont
  {Zhang}}, \bibinfo {author} {\bibfnamefont {T.}~\bibnamefont {Wang}},
  \bibinfo {author} {\bibfnamefont {Z.}~\bibnamefont {Wang}}, \bibinfo {author}
  {\bibfnamefont {B.}~\bibnamefont {Shao}},\ and\ \bibinfo {author}
  {\bibfnamefont {T.-Y.}\ \bibnamefont {Liu}},\ }\bibfield  {title} {\bibinfo
  {title} {Long-short-range message-passing: A physics-informed framework to
  capture non-local interaction for scalable molecular dynamics simulation},\
  }\href@noop {} {\bibfield  {journal} {\bibinfo  {journal} {arXiv preprint
  arXiv:2304.13542}\ } (\bibinfo {year} {2023})}\BibitemShut {NoStop}%
\bibitem [{\citenamefont {Abramson}\ \emph {et~al.}(2024)\citenamefont
  {Abramson}, \citenamefont {Adler}, \citenamefont {Dunger}, \citenamefont
  {Evans}, \citenamefont {Green}, \citenamefont {Pritzel}, \citenamefont
  {Ronneberger}, \citenamefont {Willmore}, \citenamefont {Ballard},
  \citenamefont {Bambrick} \emph {et~al.}}]{abramson2024accurate}%
  \BibitemOpen
  \bibfield  {author} {\bibinfo {author} {\bibfnamefont {J.}~\bibnamefont
  {Abramson}}, \bibinfo {author} {\bibfnamefont {J.}~\bibnamefont {Adler}},
  \bibinfo {author} {\bibfnamefont {J.}~\bibnamefont {Dunger}}, \bibinfo
  {author} {\bibfnamefont {R.}~\bibnamefont {Evans}}, \bibinfo {author}
  {\bibfnamefont {T.}~\bibnamefont {Green}}, \bibinfo {author} {\bibfnamefont
  {A.}~\bibnamefont {Pritzel}}, \bibinfo {author} {\bibfnamefont
  {O.}~\bibnamefont {Ronneberger}}, \bibinfo {author} {\bibfnamefont
  {L.}~\bibnamefont {Willmore}}, \bibinfo {author} {\bibfnamefont {A.~J.}\
  \bibnamefont {Ballard}}, \bibinfo {author} {\bibfnamefont {J.}~\bibnamefont
  {Bambrick}}, \emph {et~al.},\ }\bibfield  {title} {\bibinfo {title} {Accurate
  structure prediction of biomolecular interactions with {AlphaFold} 3},\
  }\href@noop {} {\bibfield  {journal} {\bibinfo  {journal} {Nature}\ ,\
  \bibinfo {pages} {1}} (\bibinfo {year} {2024})}\BibitemShut {NoStop}%
\bibitem [{\citenamefont {Rives}\ \emph {et~al.}(2019)\citenamefont {Rives},
  \citenamefont {Meier}, \citenamefont {Sercu}, \citenamefont {Goyal},
  \citenamefont {Lin}, \citenamefont {Liu}, \citenamefont {Guo}, \citenamefont
  {Ott}, \citenamefont {Zitnick}, \citenamefont {Ma},\ and\ \citenamefont
  {Fergus}}]{rives2019biological}%
  \BibitemOpen
  \bibfield  {author} {\bibinfo {author} {\bibfnamefont {A.}~\bibnamefont
  {Rives}}, \bibinfo {author} {\bibfnamefont {J.}~\bibnamefont {Meier}},
  \bibinfo {author} {\bibfnamefont {T.}~\bibnamefont {Sercu}}, \bibinfo
  {author} {\bibfnamefont {S.}~\bibnamefont {Goyal}}, \bibinfo {author}
  {\bibfnamefont {Z.}~\bibnamefont {Lin}}, \bibinfo {author} {\bibfnamefont
  {J.}~\bibnamefont {Liu}}, \bibinfo {author} {\bibfnamefont {D.}~\bibnamefont
  {Guo}}, \bibinfo {author} {\bibfnamefont {M.}~\bibnamefont {Ott}}, \bibinfo
  {author} {\bibfnamefont {C.~L.}\ \bibnamefont {Zitnick}}, \bibinfo {author}
  {\bibfnamefont {J.}~\bibnamefont {Ma}},\ and\ \bibinfo {author}
  {\bibfnamefont {R.}~\bibnamefont {Fergus}},\ }\bibfield  {title} {\bibinfo
  {title} {Biological structure and function emerge from scaling unsupervised
  learning to 250 million protein sequences},\ }\bibfield  {journal} {\bibinfo
  {journal} {PNAS}\ }\href {https://doi.org/10.1101/622803} {10.1101/622803}
  (\bibinfo {year} {2019})\BibitemShut {NoStop}%
\bibitem [{\citenamefont {Lin}\ \emph {et~al.}(2023)\citenamefont {Lin},
  \citenamefont {Akin}, \citenamefont {Rao}, \citenamefont {Hie}, \citenamefont
  {Zhu}, \citenamefont {Lu}, \citenamefont {Smetanin}, \citenamefont {Verkuil},
  \citenamefont {Kabeli}, \citenamefont {Shmueli} \emph
  {et~al.}}]{lin2023evolutionary}%
  \BibitemOpen
  \bibfield  {author} {\bibinfo {author} {\bibfnamefont {Z.}~\bibnamefont
  {Lin}}, \bibinfo {author} {\bibfnamefont {H.}~\bibnamefont {Akin}}, \bibinfo
  {author} {\bibfnamefont {R.}~\bibnamefont {Rao}}, \bibinfo {author}
  {\bibfnamefont {B.}~\bibnamefont {Hie}}, \bibinfo {author} {\bibfnamefont
  {Z.}~\bibnamefont {Zhu}}, \bibinfo {author} {\bibfnamefont {W.}~\bibnamefont
  {Lu}}, \bibinfo {author} {\bibfnamefont {N.}~\bibnamefont {Smetanin}},
  \bibinfo {author} {\bibfnamefont {R.}~\bibnamefont {Verkuil}}, \bibinfo
  {author} {\bibfnamefont {O.}~\bibnamefont {Kabeli}}, \bibinfo {author}
  {\bibfnamefont {Y.}~\bibnamefont {Shmueli}}, \emph {et~al.},\ }\bibfield
  {title} {\bibinfo {title} {Evolutionary-scale prediction of atomic-level
  protein structure with a language model},\ }\href@noop {} {\bibfield
  {journal} {\bibinfo  {journal} {Science}\ }\textbf {\bibinfo {volume}
  {379}},\ \bibinfo {pages} {1123} (\bibinfo {year} {2023})}\BibitemShut
  {NoStop}%
\bibitem [{\citenamefont {Mardt}\ \emph {et~al.}(2022)\citenamefont {Mardt},
  \citenamefont {Hempel}, \citenamefont {Clementi},\ and\ \citenamefont
  {No{\'e}}}]{mardt2022deep}%
  \BibitemOpen
  \bibfield  {author} {\bibinfo {author} {\bibfnamefont {A.}~\bibnamefont
  {Mardt}}, \bibinfo {author} {\bibfnamefont {T.}~\bibnamefont {Hempel}},
  \bibinfo {author} {\bibfnamefont {C.}~\bibnamefont {Clementi}},\ and\
  \bibinfo {author} {\bibfnamefont {F.}~\bibnamefont {No{\'e}}},\ }\bibfield
  {title} {\bibinfo {title} {Deep learning to decompose macromolecules into
  independent markovian domains},\ }\href@noop {} {\bibfield  {journal}
  {\bibinfo  {journal} {Nature Communications}\ }\textbf {\bibinfo {volume}
  {13}},\ \bibinfo {pages} {7101} (\bibinfo {year} {2022})}\BibitemShut
  {NoStop}%
\bibitem [{\citenamefont {Nakata}\ and\ \citenamefont
  {Shimazaki}(2017)}]{nakata2017pubchemqc}%
  \BibitemOpen
  \bibfield  {author} {\bibinfo {author} {\bibfnamefont {M.}~\bibnamefont
  {Nakata}}\ and\ \bibinfo {author} {\bibfnamefont {T.}~\bibnamefont
  {Shimazaki}},\ }\bibfield  {title} {\bibinfo {title} {{PubChemQC} project: a
  large-scale first-principles electronic structure database for data-driven
  chemistry},\ }\href@noop {} {\bibfield  {journal} {\bibinfo  {journal}
  {Journal of Chemical Information and Modeling}\ }\textbf {\bibinfo {volume}
  {57}},\ \bibinfo {pages} {1300} (\bibinfo {year} {2017})}\BibitemShut
  {NoStop}%
\bibitem [{\citenamefont {Simeon}\ and\ \citenamefont
  {De~Fabritiis}(2023)}]{simeon2023tensornet}%
  \BibitemOpen
  \bibfield  {author} {\bibinfo {author} {\bibfnamefont {G.}~\bibnamefont
  {Simeon}}\ and\ \bibinfo {author} {\bibfnamefont {G.}~\bibnamefont
  {De~Fabritiis}},\ }\bibfield  {title} {\bibinfo {title} {Tensornet: Cartesian
  tensor representations for efficient learning of molecular potentials},\
  }\href@noop {} {\bibfield  {journal} {\bibinfo  {journal} {Advances in Neural
  Information Processing Systems}\ }\textbf {\bibinfo {volume} {36}},\ \bibinfo
  {pages} {37334} (\bibinfo {year} {2023})}\BibitemShut {NoStop}%
\bibitem [{\citenamefont {Everett}\ \emph {et~al.}(2024)\citenamefont
  {Everett}, \citenamefont {Xiao}, \citenamefont {Wortsman}, \citenamefont
  {Alemi}, \citenamefont {Novak}, \citenamefont {Liu}, \citenamefont {Gur},
  \citenamefont {{Sohl-Dickstein}}, \citenamefont {Kaelbling}, \citenamefont
  {Lee},\ and\ \citenamefont {Pennington}}]{everett_scaling_2024}%
  \BibitemOpen
  \bibfield  {author} {\bibinfo {author} {\bibfnamefont {K.}~\bibnamefont
  {Everett}}, \bibinfo {author} {\bibfnamefont {L.}~\bibnamefont {Xiao}},
  \bibinfo {author} {\bibfnamefont {M.}~\bibnamefont {Wortsman}}, \bibinfo
  {author} {\bibfnamefont {A.~A.}\ \bibnamefont {Alemi}}, \bibinfo {author}
  {\bibfnamefont {R.}~\bibnamefont {Novak}}, \bibinfo {author} {\bibfnamefont
  {P.~J.}\ \bibnamefont {Liu}}, \bibinfo {author} {\bibfnamefont
  {I.}~\bibnamefont {Gur}}, \bibinfo {author} {\bibfnamefont {J.}~\bibnamefont
  {{Sohl-Dickstein}}}, \bibinfo {author} {\bibfnamefont {L.~P.}\ \bibnamefont
  {Kaelbling}}, \bibinfo {author} {\bibfnamefont {J.}~\bibnamefont {Lee}},\
  and\ \bibinfo {author} {\bibfnamefont {J.}~\bibnamefont {Pennington}},\
  }\bibfield  {title} {\bibinfo {title} {Scaling exponents across
  parameterizations and optimizers},\ }\href@noop {} {\bibfield  {journal}
  {\bibinfo  {journal} {arXiv preprint arXiv:2407.05872}\ } (\bibinfo {year}
  {2024})}\BibitemShut {NoStop}%
\bibitem [{\citenamefont {Hempel}\ \emph {et~al.}(2020)\citenamefont {Hempel},
  \citenamefont {Plattner},\ and\ \citenamefont
  {No{\'e}}}]{hempel_coupling_2020}%
  \BibitemOpen
  \bibfield  {author} {\bibinfo {author} {\bibfnamefont {T.}~\bibnamefont
  {Hempel}}, \bibinfo {author} {\bibfnamefont {N.}~\bibnamefont {Plattner}},\
  and\ \bibinfo {author} {\bibfnamefont {F.}~\bibnamefont {No{\'e}}},\
  }\bibfield  {title} {\bibinfo {title} {Coupling of conformational switches in
  calcium sensor unraveled with local {{Markov}} models and transfer entropy},\
  }\href@noop {} {\bibfield  {journal} {\bibinfo  {journal} {Journal of
  Chemical Theory and Computation}\ }\textbf {\bibinfo {volume} {16}},\
  \bibinfo {pages} {2584} (\bibinfo {year} {2020})}\BibitemShut {NoStop}%
\bibitem [{\citenamefont {Hempel}\ \emph {et~al.}(2022)\citenamefont {Hempel},
  \citenamefont {Plattner},\ and\ \citenamefont {Noe}}]{hempel_molecular_2022}%
  \BibitemOpen
  \bibfield  {author} {\bibinfo {author} {\bibfnamefont {T.}~\bibnamefont
  {Hempel}}, \bibinfo {author} {\bibfnamefont {N.}~\bibnamefont {Plattner}},\
  and\ \bibinfo {author} {\bibfnamefont {F.}~\bibnamefont {Noe}},\ }\bibfield
  {title} {\bibinfo {title} {Molecular dynamics dataset of synaptotagmin-1},\
  }\href {https://doi.org/10.5281/zenodo.6908073} {10.5281/zenodo.6908073}
  (\bibinfo {year} {2022})\BibitemShut {NoStop}%
\end{thebibliography}
%apsrev4-2.bst 2019-01-14 (MD) hand-edited version of apsrev4-1.bst
%Control: key (0)
%Control: author (8) initials jnrlst
%Control: editor formatted (1) identically to author
%Control: production of article title (0) allowed
%Control: page (0) single
%Control: year (1) truncated
%Control: production of eprint (0) enabled
%

\appendix

\setcounter{figure}{0} % reset figure counter for Supp. Figures
\setcounter{equation}{0} % reset equation counter for Supp. Equations
\makeatletter 
\renewcommand{\thefigure}{S\@arabic\c@figure} % make Figure legend start with Figure S
\renewcommand{\thetable}{S\@arabic\c@table} 
\makeatother
\def\theequation{S\arabic{equation}}

\clearpage
\section{Training details and hyperparameters}\label{appendix:training}

For pretraining, we use the PCQM4Mv2 dataset\cite{nakata2017pubchemqc}, which consists of 3.7 million three-dimensional conformations of small-size molecules. These equilibrium conformations are obtained with calculations of the B3LYP / 6-31G$^*$ level of density functional theory. For pretraining, we randomly take 10,000 configurations as the validation set and the remainder as the training set. 

\begin{table}[htb]
    \centering
    \begin{minipage}[t]{0.45\textwidth}
        \centering
        \caption{Hyperparameters for the pretrained network.}
        \vspace{0pt} % Ensures alignment at the top
        \begin{tabular}{c|c}
            \textbf{Hyperparameter} & \textbf{Value} \\
            \hline
            Architecture & TensorNet\cite{simeon2023tensornet}\\
            $d$    & 64 \\
            \# of MP layers     & 3 \\
            Batch size          & 100 \\
            Epochs              & 10 \\
            \# of RBFs           & 32 \\
            $r_{\text{cut}}$ ($\text{\AA}$)     & 5 \\
            Learning rate       & 0.0005 \\
            Optimizer           & AdamW (AMSGrad) \\
            Noise level         & 0.05 \\ \hline
        \end{tabular}
        \label{tab:hp_pretrain}
    \end{minipage}
\end{table}

\begin{table}[htb]
\centering
\caption{Hyperparameters for VAMPnets}
\begin{tabular}{c|ccc}
\textbf{Hyperparameter} & \textbf{C2A} & \textbf{ADK} & \textbf{nsp13-ADP}\\ \hline
Batch size & \multicolumn{3}{c}{1000}\\
Learning rate & \multicolumn{3}{c}{0.0005}\\
Optimizer & \multicolumn{3}{c}{AdamAtan2\cite{everett_scaling_2024}} \\
Maximum epochs & \multicolumn{3}{c}{5} \\
Training patience & \multicolumn{3}{c}{1000} \\
Validation patience & \multicolumn{3}{c}{10} \\
Validation interval & \multicolumn{3}{c}{50} \\
Hidden dimension & 64 & 64 &128 \\
MLP activation & \multicolumn{3}{c}{SiLU} \\
\# of MLP layers & \multicolumn{3}{c}{2} \\
Trajectory stride & 10 & 10 & 1\\
Training lag time (ns) & 0.05 & 4 & 12\\
Dropout & \multicolumn{3}{c}{0.1}\\
Output dimension ($k$) & 20 & 30 & 4 \\
% Architectures & \multicolumn{3}{c}{SF-Merger(Merging-only/GCN/GC/RGGC/TAG)}\\
\hline
\end{tabular}
\label{tab:hp_vamp}
\end{table}

\begin{table}[htb]
    \begin{minipage}[t]{0.45\textwidth}
        \centering
        \caption{Hyperparameters for SPIB}
        \vspace{0pt} % Ensures alignment at the top
        \begin{tabular}{c|cc}
            \textbf{Hyperparameter}  & \textbf{ADK} & \textbf{nsp13-ADP} \\ \hline
            Batch size & \multicolumn{2}{c}{1000} \\
            Learning rate & \multicolumn{2}{c}{0.0002} \\
            Optimizer & \multicolumn{2}{c}{AdamAtan2} \\
            Label refine iteration. & \multicolumn{2}{c}{5} \\
            Label refine patience & \multicolumn{2}{c}{10} \\
            Hidden dimension & \multicolumn{2}{c}{64} \\
            MLP activation & \multicolumn{2}{c}{SiLU} \\
            \# of MLP layers & \multicolumn{2}{c}{2} \\
            Trajectory stride & 20 & 1 \\
            Training lag time (ns) & 10 & 30 \\
            $d$ & \multicolumn{2}{c}{64} \\
            Dropout & \multicolumn{2}{c}{0.1} \\
            Label refinement frequency & \multicolumn{2}{c}{5} \\
            Architecture & RGGC & GC \\
            \# of transformer layers & \multicolumn{2}{c}{3} \\
            Window size & \multicolumn{2}{c}{6} \\ 
            \hline
        \end{tabular}
        \label{tab:hp_spib}
    \end{minipage} 
\end{table}

\section{Local message-passing operators for the fragment graph}
\label{app:local-mp}

%\subsection{Notation and setup}

Each node $i$ in the fragment graph has neighbors $\mathcal{N}(i)$ and feature vector $\mathbf{x}_i \in \mathbb{R}^H$. We denote the adjacency and degree matrices by $\mathbf{A}$ and $\mathbf{D}$ and the corresponding quantities when self-loops are included by $\tilde{\mathbf{A}}=\mathbf{A}+\mathbf{I}$ and $\tilde{\mathbf{D}}=\mathbf{D}+\mathbf{I}$. The graph operators that we consider for the TMM graph-convolution leayer are all single-layer message-passing updates of the form
\begin{equation}
\mathbf{x}'_i \;=\; \mathbf{W}_0\,\mathbf{x}_i \;+\; \sum_{j\in\mathcal{N}(i)} \alpha_{ij}\,\mathbf{W}_1\,\mathbf{x}_j,
\end{equation}
with operator-specific weights $\alpha_{ij}$ and trainable matrices $\{\mathbf{W}_k\}_{k=0}^{K}$.  We describe each in turn.

%\subsection{Operator definitions}
\subsection{GCNConv (GCN)~\cite{kipf2016semi}.}
\begin{equation}
\label{eq:A1}
\mathbf{x}'_i \;=\; \mathbf{W}_0\,\frac{1}{\tilde d_i}\mathbf{x}_i \;+\;
\mathbf{W}_1 \sum_{j\in\mathcal{N}(i)} \frac{1}{\sqrt{\tilde d_i\,\tilde d_j}}\,\mathbf{x}_j,
\tag{A1}
\end{equation}
where $\tilde d_i = \sum_j D_{ij}$.

\subsection{GraphConv (GC)~\cite{morris2019weisfeiler}.}
\begin{equation}
\label{eq:A2}
\mathbf{x}'_i \;=\; \mathbf{W}_0\,\mathbf{x}_i \;+\; \mathbf{W}_1 \sum_{j\in\mathcal{N}(i)} \mathbf{x}_j.
\tag{A2}
\end{equation}

\subsection{ResGatedGraphConv (RGGC)~\cite{bresson2017residual}.}
\begin{align}
\label{eq:A3}
\mathbf{x}'_i &= \mathbf{W}_0\,\mathbf{x}_i \;+\;
\sum_{j\in\mathcal{N}(i)} \Bigl[\eta_{ij} \odot \mathbf{W}_1\,\mathbf{x}_j\Bigr], \tag{A3}\\
%\eta_{ij} &= \sigma\!\bigl(\mathbf{W}_2\,\mathbf{x}_i + \mathbf{W}_3\,\mathbf{x}_j\bigr),
\end{align}
where $\eta_{ij}= \sigma(\mathbf{W}_2\,\mathbf{x}_i + \mathbf{W}_3\,\mathbf{x}_j)$ gates neighbor contributions elementwise, and $\sigma$ is a sigmoid function.

\subsection{TAGConv~\cite{du2017topology}.}
\begin{align}
\label{eq:A4}
\mathbf{x}'_i \;&=\; \sum_{k=0}^{K} \mathbf{W}_k \sum_{j\in\mathcal{N}(i)} [\mathbf{T}^{\,k}]_{ij}\,\mathbf{x}_j, \tag{A4}\\
%\mathbf{T} \;&=\; \mathbf{D}^{-\tfrac{1}{2}}\mathbf{A}\mathbf{D}^{-\tfrac{1}{2}}.
\end{align}
where $\mathbf{T}=\mathbf{D}^{-1/2}\mathbf{A}\mathbf{D}^{-1/2}$ is the symmetrically normalized matrix.

%\subsection{Summary and implementation details}
Table~\ref{tab:mp-summary} summarizes key differences. All four operators scale as $\mathcal{O}(|E|H)$, where $E$ is the set of edges, and we use a single layer in the token-merging module to keep the cost linear in the number of fragments. We use the PyTorch Geometric~\cite{Fey/Lenssen/2019} implementations with default hyperparameters unless otherwise stated. Hyperparameters such as cutoff or basis sizes are listed in Table~\ref{tab:hp_vamp} and ~\ref{tab:hp_spib}.

\begin{table*}
\centering
\caption{Local message-passing operators used in the fragment graph.}
\label{tab:mp-summary}
\begin{tabular}{lcccc}
\toprule
Operator & Self-loop & Normalization & Special weighting & PyG class \\
\midrule
GCNConv & yes ($\tilde{\mathbf{A}}$) & degree, symmetric & --- & \texttt{GCNConv} \\
GraphConv (GC) & optional & none & --- & \texttt{GraphConv} \\
ResGatedGraphConv (RGGC) & optional & none & gate $\eta_{ij}$ & \texttt{ResGatedGraphConv} \\
TAGConv & yes (via $\mathbf{T}$) & degree, symmetric & $K$-hop polynomial & \texttt{TAGConv} \\
\bottomrule
\end{tabular}
\end{table*}

\section{Additional results on synaptogamin C2A}
\label{sec:c2a}

Synaptotagmin C2A is the calcium binding domain involved in synaptic exocytosis whose conformational dynamics in the presence and absence of calcium have been studied computationally~\cite{hempel_coupling_2020}. This system has been used as a reference in iVAMPnet\cite{mardt2022deep}, where the main movement can be described by collective movements in the calcium binding region and loops in C34 and C78. Because there is a lack of a clear conformational change, we chose to use this system as a reference rather than elaborating the dimensionality reduction results. We use the dataset from \citet{hempel_coupling_2020} which contains a total of 184 $\mu$s of simulation time\cite{hempel_molecular_2022}.

%\section{Supplementary Figures}

\begin{figure*}
    \centering
    \includegraphics[width=\textwidth]{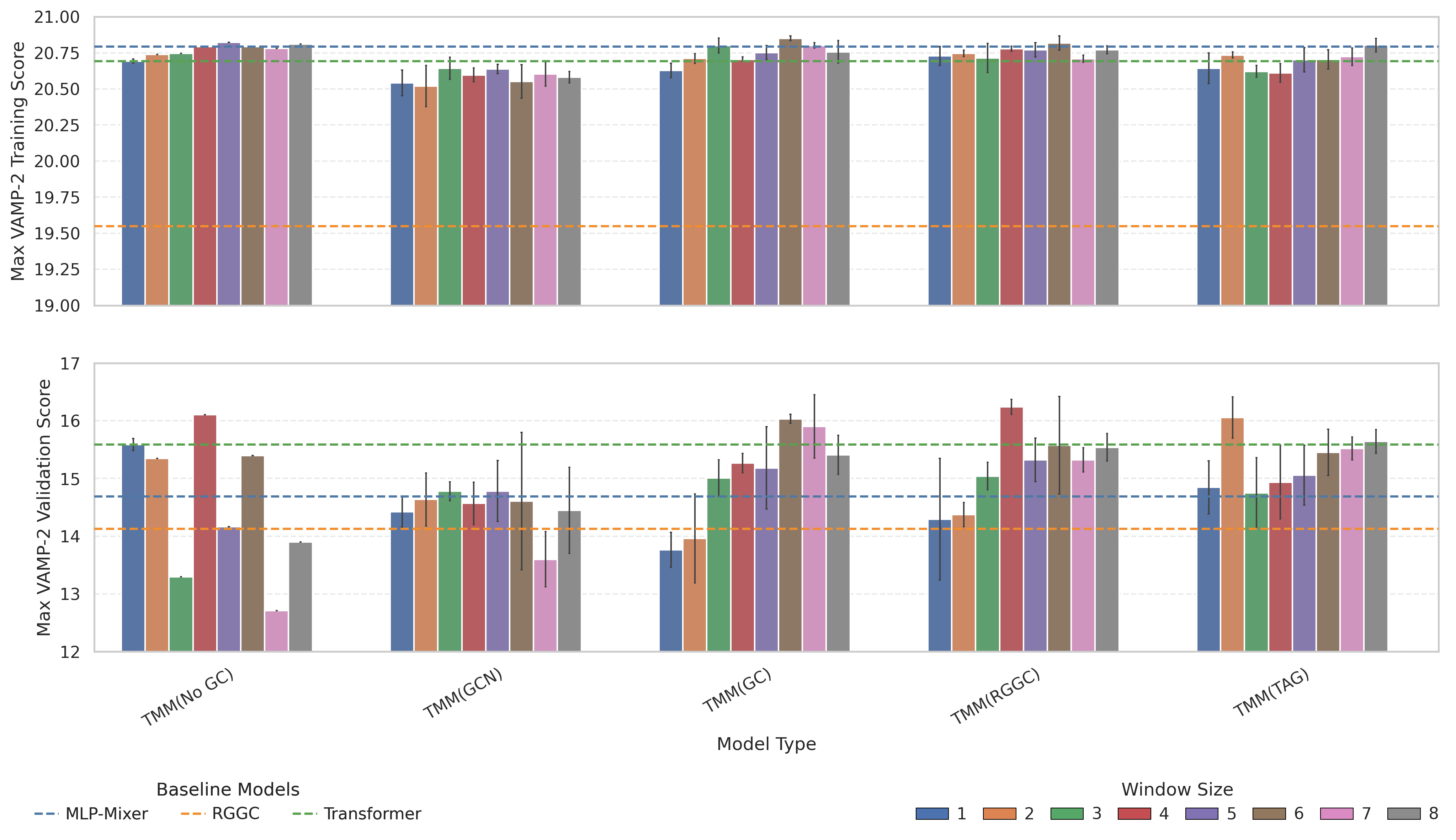}
    \caption{Maximum training (top) and validation (bottom) VAMP-2 scores for synaptotagmin C2A from using various architectures. Error bars show standard deviation over three runs. A window size of 1 is equivalent to no token merging. The training and validation data are held fixed across repeated runs and models. ``RGGC'' is a GNN baseline consisting of 4 RGGC layers with mean pooling (no mixer).}
    \label{fig:scores_c2a}
\end{figure*}

\begin{figure*}
    \centering
    \includegraphics{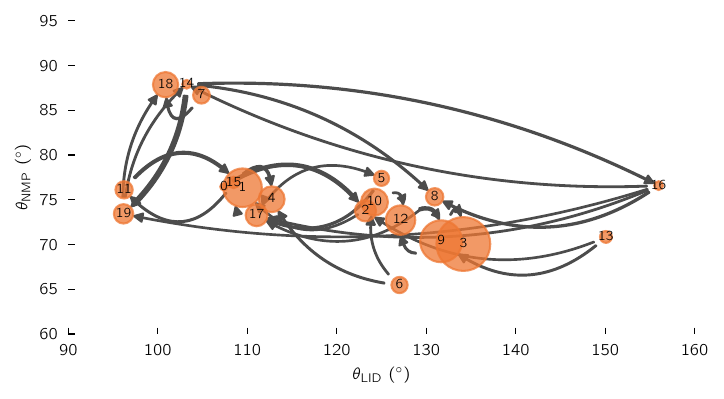}
    \caption{MSM of SPIB-learned states for ADK estimated with a lag time of 10~ns. Each labeled node represents a metastable state, and the size of each node is proportional to its population.  The edges indicate the highest-probability transitions between states, with thicker edges denoting a larger number of transitions per unit time. The states are plotted according to the average NMP-CORE and LID-CORE angles, $\theta_\mathrm{NMP}$ and $\theta_\mathrm{LID}$. The NMP-CORE angle $\theta_\mathrm{NMP}$ is defined by the centers of geometry of the backbone and $\ce{C_\beta}$ atoms in residue groups 115--125, 90--100, and 35--55. The LID-CORE angle $\theta_\mathrm{LID}$ is defined similarly using residue groups 179--185, 115--125, and 125--153~\cite{beckstein_zipping_2009}.}
    \label{fig:spib_network_adk_theta}
\end{figure*}

\begin{figure*}
    \centering
    \includegraphics[width=\linewidth]{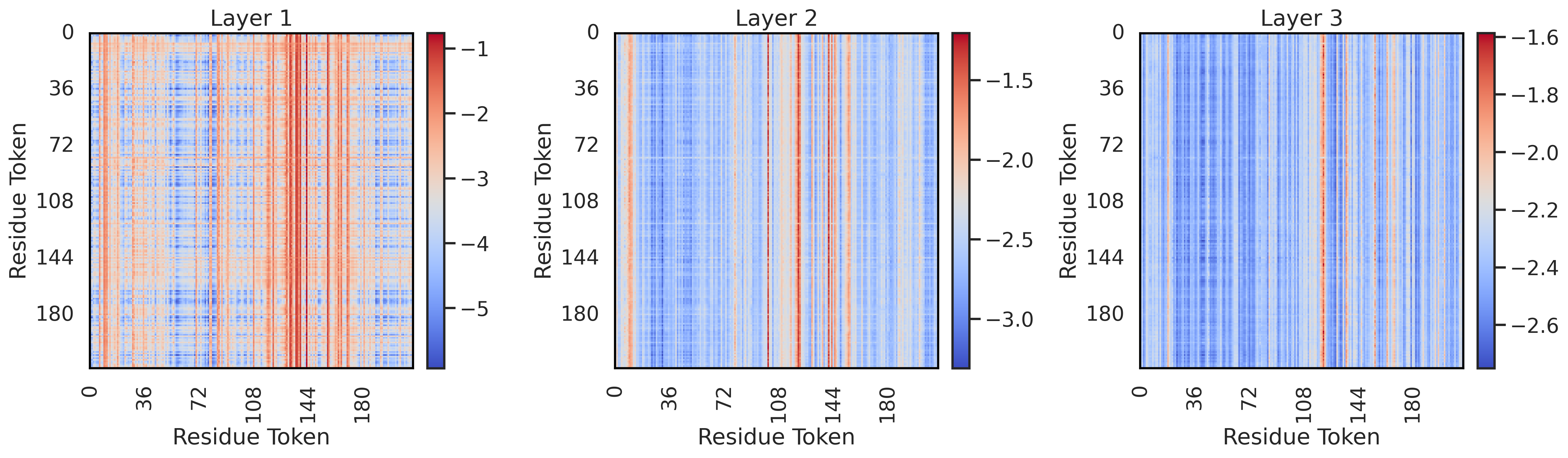}
    \caption{Maps of log-attention weights from each layer of a transformer (without TMM) averaged across all SPIB conformational states and across attention heads. Each index is a residue token. }
    \label{fig:spib_plain_transformer_attn}
\end{figure*}

\end{document}